\documentclass[10pt,twocolumn,letterpaper]{article}
\newcommand{\xmark}{\ding{55}}%

\newcommand{\orcid}[1]{\href{https://orcid.org/#1}{\textcolor[HTML]{A6CE39}{\aiOrcid}}}

\usepackage[algorithms]{wacv}      % To produce the REVIEW version for the algorithms track
\definecolor{wacvblue}{rgb}{0.21,0.49,0.74}
\usepackage[pagebackref,breaklinks,colorlinks,allcolors=wacvblue]{hyperref}

\def\wacvPaperID{2046} % *** Enter the WACV Paper ID here
\def\confName{WACV}
\def\confYear{2026}
\usepackage{multirow}
\usepackage{comment}
\usepackage{arydshln}
\usepackage{xcolor}
\usepackage{multicol}
\usepackage{pifont}
\usepackage{float}
\title{DexAvatar: 3D Sign Language Reconstruction with Hand and Body Pose Priors}

\author{
Kaustubh Kundu$^{1}$,
Hrishav Bakul Barua$^{1,2}$,
Lucy Robertson-Bell$^{1}$,
Zhixi Cai$^{1}$,
Kalin Stefanov$^{1}$ \\
$^{1}$Monash University \\
$^{2}$TCS Research \\
{\tt\small \{kaustubh.kundu, hrishav.barua, lucy.robertson-bell, zhixi.cai, kalin.stefanov\}@monash.edu}
}

\newcommand{\mymethod}{DexAvatar}
\begin{document}
% \maketitle

\twocolumn[{
\maketitle
\begin{center}
\captionsetup{type=figure}
\includegraphics[width=\textwidth]{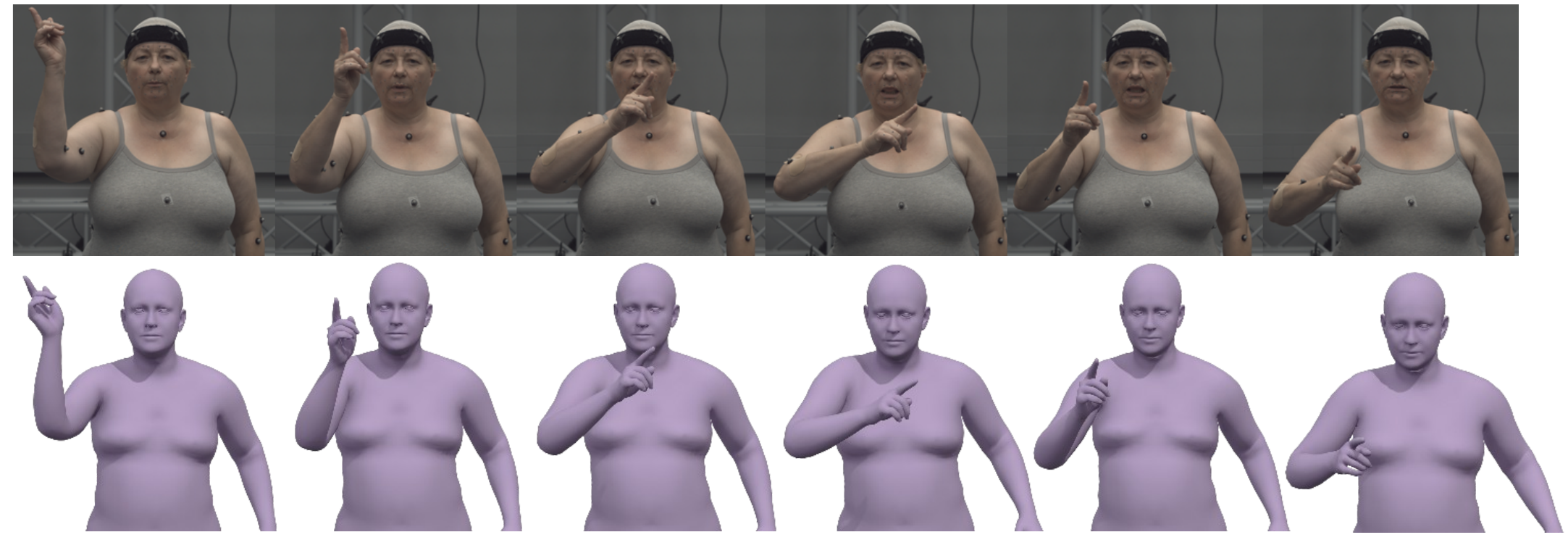}
\captionof{figure}{\textbf{DexAvatar} recovers bio-mechanically accurate 3D hand and body poses from monocular sign language videos.
}
\label{qualitative_results_healthy}
\end{center}
}]

% \begin{figure*}[t]
%     \centering
%     \includegraphics[width=\textwidth]{wacv-2026-author-kit-template/Figures/Intro_diagram.pdf}
%     \caption{Visualization of results for T2w and T1w Healthy brain MRI synthesis with corresponding error maps. McCaD yields lower artifacts with higher anatomical fidelity compared to baselines.}
%     \label{qualitative_results_healthy}
% \end{figure*}

\begin{abstract}
The trend in sign language generation is centered around data-driven generative methods that require vast amounts of precise 2D and 3D human pose data to achieve an acceptable generation quality.
However, currently, most sign language datasets are video-based and limited to automatically reconstructed 2D human poses (i.e., keypoints) and lack accurate 3D information.
% However, manual production of accurate 2D and 3D human pose information from videos is a labor-intensive process.
Furthermore, existing state-of-the-art for automatic 3D human pose estimation from sign language videos is prone to self-occlusion, noise, and motion blur effects, resulting in poor reconstruction quality.
In response to this, we introduce DexAvatar, a novel framework to reconstruct bio-mechanically accurate fine-grained hand articulations and body movements from in-the-wild monocular sign language videos, guided by learned 3D hand and body priors.
DexAvatar achieves strong performance in the SGNify motion capture dataset, the only benchmark available for this task, reaching an improvement of 35.11\% in the estimation of body and hand poses compared to the state-of-the-art. The official website of this work is: \textcolor{red}{https://github.com/kaustesseract/DexAvatar}.
\end{abstract}    
\section{Introduction}
Sign languages are the primary mode of communication for approximately 466 million Deaf or hard-of-hearing individuals worldwide~\cite{who}.
They are spatio-temporal languages that utilize the visual–gestural modality to convey meaning through manual hand articulations in combination with non-manual elements like the face and body.
Similarly to spoken languages, sign languages follow linguistic rules~\cite{signling}, but lack standardized written forms \eg, American Sign Language is not a visual representation of spoken English.

Existing sign language datasets~\cite{lsa,psl,wlasl,autsl,italian,csl} provide videos and 2D keypoints and have contributed significantly to sign language generation~\cite{saunders2020progressive,saunders2021mixed,arkushin2023ham2pose}.
%However, they do not capture the full spatial complexity of signing, especially depth and contact, and different 3D configurations can project to the same 2D keypoints.
However, 2D representations cannot capture the full spatial complexity of signing. Depth information and hand-body contact are critical for conveying meaning, yet different 3D hand configurations can project to identical 2D keypoints.
%There is a clear need for datasets with accurate 3D information to support realistic spatial sign language generation.
There is therefore a clear need for datasets with accurate 3D information to support realistic spatial sign language generation.
%With the advent of advanced whole-body parametric models, such as SMPL-X~\cite{smplx}, 3D reconstruction of human pose and shape has gained significant attention.
%Expressive whole-body mesh recovery jointly estimates 3D body and hand pose, and facial expressions from videos.
Recent advances in whole-body parametric models, such as SMPL-X~\cite{smplx}, enable expressive whole-body mesh recovery that jointly estimates 3D body and hand pose and facial expressions from videos.

%Despite significant progress in whole-body mesh recovery, 3D hand pose estimation remains challenging for sign language videos.
Despite significant progress, 3D hand pose estimation remains challenging for sign language videos.
%Hand movements in sign language are more complex and intricate compared to everyday scenarios; the main challenges arise from the complex articulations of the hands~\cite{bilal2011vision}, as well as frequent hand-to-hand and hand-to-body interactions leading to occlusions~\cite{moon2020interhand2}.
Hand movements in sign language are more complex and intricate compared to everyday scenarios. The main challenges arise from rapid and complex hand articulations~\cite{bilal2011vision}, frequent hand-to-hand and hand-to-body interactions that cause self-occlusions~\cite{moon2020interhand2}, and motion blur from fast signing motions.
%Furthermore, low-resolution and motion blur artifacts caused by rapid sign motions often hinder the accurate hand pose reconstruction across many video frames.
%As a result, existing whole-body and hand-only mesh recovery methods struggle to estimate reliable poses from sign language videos.
These factors hinder accurate hand pose reconstruction across many video frames, and existing whole-body and hand-only mesh recovery methods struggle to estimate reliable poses from sign language videos.

%There are two approaches to whole-body mesh recovery: regression-based prediction and optimization-based fitting.
Current approaches to whole-body mesh recovery fall into two categories: regression-based and optimization-based methods.
%Regression methods predict mesh parameters directly and are efficient~\cite{lin2023one,cai2023smpler,sun2024aios,li2025unipose}, but they are usually trained on general-purpose data and often do not capture hand articulations specific to signing.
Regression methods predict mesh parameters directly and are efficient~\cite{lin2023one,cai2023smpler,sun2024aios,li2025unipose}, but are usually trained on general-purpose data and often fail to capture the hand articulations specific to signing.
Optimization methods fit parametric models using priors and multi-term objectives and are more computationally intensive, but can incorporate sign language aware constraints to yield accurate and stable hand poses under self-occlusion.
We therefore introduce \textbf{DexAvatar}, an optimization framework that incorporates signing-based hand and body priors to reconstruct 3D signing avatars from monocular videos.

%Existing work~\cite{smplx, baltatzis2024neural} trained priors using general-purpose datasets that often fail to capture the true nature of signed communication.
A key limitation of existing work~\cite{smplx, baltatzis2024neural} is that priors are trained on general-purpose datasets that fail to capture the distinctive characteristics of signed communication.
%In sign languages, meaning relies heavily on hand shape and orientation, however, accurate 3D hand data are not publicly available.
In sign languages, meaning relies heavily on precise hand shape and orientation, yet accurate 3D hand data are not publicly available.
%Therefore, we collected a motion capture sign language dataset to capture fine finger articulations and train the proposed hand prior, SignHPoser.
To address this, we collected a motion capture sign language dataset to capture fine finger articulations and train our proposed hand prior, SignHPoser.
The proposed body prior, SignBPoser, is trained with a subset of the 3D data published in \cite{yu2024signavatars}, reconstructed from the How2Sign~\cite{how2sign} dataset.
The pretrained body and hand priors can be utilized in any regression- or optimization-based approach for whole-body mesh recovery from sign language videos.
Our main contributions are:

\begin{itemize}
    \item We introduce two sign language-aware pose priors, SignHPoser for hands and SignBPoser for body, trained to learn compact latent spaces that preserve phonologically meaningful variations and discourage anatomically implausible configurations.
    \item We integrate the priors into DexAvatar, an optimization pipeline that reconstructs 3D signing avatars from monocular videos. DexAvatar
    employs the priors as differentiable regularizers, together with temporal consistency and contact-aware terms, to stabilize estimation under self-occlusion and noisy 2D keypoints in upper-body-only and one-handed signing videos.
    \item Extensive experiments show that DexAvatar consistently achieves lower reconstruction errors compared to strong baselines for whole-body and hand-only mesh recovery.
\end{itemize}
\section{Related Work}
A foundational element of our ability to interact with others is the recognition of body poses.
Similarly, for intelligent machines, the accurate understanding of body poses is essential for interacting with humans.

\noindent\textbf{2D Human Pose Estimation.}
Research on 2D keypoints estimation has evolved from early tree- and random forest-based models~\cite{er1,er2,er3} to recent deep learning methods~\cite{tsd,vitpose,visiontrans,khirodkar2024sapiens}.
%Recently, 3D joint estimation~\cite{tang20233d,zhu2023motionbert,foo2023unified,liu2025tcpformer} methods have shown promise in overcoming the depth ambiguity present in 2D keypoints methods.
Recently, 3D joint estimation methods~\cite{tang20233d,zhu2023motionbert,foo2023unified,liu2025tcpformer} have shown promise in overcoming the depth ambiguity inherent to 2D approaches.
However, these approaches predict a sparse set of skeletal keypoints, limiting expressivity.

\noindent\textbf{3D Human Pose Reconstruction.}
%Prominent approaches for expressive 3D body models such as SMPL-X~\cite{smplx}, MANO~\cite{romero2022embodied}, and FLAME~\cite{li2017learning}, have enabled research on estimating 3D body meshes.
Expressive 3D body models such as SMPL-X~\cite{smplx}, MANO~\cite{romero2022embodied}, and FLAME~\cite{li2017learning} have enabled research on estimating 3D body meshes.
%For holistic 3D mesh recovery, prior work leverages multi-task learning~\cite{li2025unipose}, hybrid priors~\cite{sun2024aios}, and attention mechanisms~\cite{cai2023smpler,lin2023one}. To recover specific body parts such as hands, prior work often incorporates auxiliary components for hand localization and bounding box refinement~\cite{potamias2025wilor,zhang2025hawor}, or integrates optimization-based refinement and temporal filtering~\cite{dong2024hamba,pavlakos2024reconstructing} to improve stability and accuracy.
For holistic 3D mesh recovery, prior work leverages multi-task learning~\cite{li2025unipose}, hybrid priors~\cite{sun2024aios}, and attention mechanisms~\cite{cai2023smpler,lin2023one}. For hand-specific recovery, methods incorporate auxiliary components for hand localization and bounding box refinement~\cite{potamias2025wilor,zhang2025hawor}, or integrate optimization-based refinement and temporal filtering~\cite{dong2024hamba,pavlakos2024reconstructing} to improve stability and accuracy.
Although these methods offer strong potential for expressive sign language reconstruction, they are generic and fail to handle long self-occlusions, enforce realistic hand-hand and hand-body contact, and capture fine finger articulations in sign language videos.

\noindent\textbf{3D Avatar Reconstruction for Sign Language.}
%SGNify~\cite{forte2023reconstructing} introduced one of the first dedicated pipelines for whole-body mesh recovery, adding linguistic priors that constrain 3D hand pose to resolve ambiguities.
SGNify~\cite{forte2023reconstructing} introduced one of the first dedicated pipelines for whole-body mesh recovery from sign language videos, adding linguistic priors that constrain 3D hand pose to resolve ambiguities.
Built atop SMPLify-X~\cite{smplx}, it estimates SMPL-X~\cite{smplx} parameters from images.
%However, reliance on pseudo-ground truth from off-the-shelf 2D keypoints detectors such as ViTPose~\cite{xu2022vitpose} and MediaPipe~\cite{lugaresi2019mediapipe} can limit accuracy.
However, reliance on pseudo-ground truth from off-the-shelf 2D keypoint detectors such as ViTPose~\cite{xu2022vitpose} and MediaPipe~\cite{lugaresi2019mediapipe} limits accuracy.
%The method also struggles in scenarios with severe self-occlusions, intricate hand–hand and hand–body interactions, motion blur, and frequent cropping, all of which are inherent to signing and pose significant challenges to accurate reconstruction.
The method also struggles with severe self-occlusions, intricate hand–hand and hand–body interactions, motion blur, and frequent cropping—all inherent challenges in sign language reconstruction.

OSX~\cite{lin2023one} proposed the UBody pipeline for downstream upper body reconstruction tasks, including sign language, gesture, and emotion generation.
%While OSX demonstrates high performance on general-purpose datasets, it often misses nuanced articulations and can produce unrealistic hand poses in sign language videos. 
While OSX demonstrates strong performance on general-purpose datasets, it often misses nuanced articulations and produces unrealistic hand poses in sign language videos.

Neural Sign Actors~\cite{baltatzis2024neural} and SignAvatars~\cite{yu2024signavatars} used OSX for initialization.
The former used OSX to initialize pose and shape for an optimization-based pipeline and added a mesh prior based on a principal component analysis-based space derived from the AMASS~\cite{mahmood2019amass} and ARCTIC~\cite{fan2023arctic} datasets.
The latter fused OSX with ACR~\cite{yu2023acr} and PARE~\cite{kocabas2021pare} and used a pseudo-ground truth 2D keypoints~\cite{lugaresi2019mediapipe, vitpose} within a SMPLify-X~\cite{smplx} framework.

%EVA~\cite{hu2024expressive} also leveraged multiple sources as pseudo-ground truth, including initial estimates of camera parameters, SMPL-X parameters~\cite{cai2023smpler}, 2D keypoints~\cite{pavlakos2024reconstructing}, and 3D hand parameters~\cite{cai2023smpler} from off-the-shelf tools such as~\cite{yang2023effective}.
EVA~\cite{hu2024expressive} leveraged multiple sources as pseudo-ground truth, including initial estimates of camera parameters, SMPL-X parameters~\cite{cai2023smpler}, 2D keypoints~\cite{pavlakos2024reconstructing}, and 3D hand parameters~\cite{cai2023smpler} from off-the-shelf tools~\cite{yang2023effective}.
%Despite the effectiveness in denoising and generating plausible motions, existing pipelines suffer from critical limitations in the context of sign languages.
Despite their effectiveness in denoising and generating plausible motions, existing pipelines suffer from critical limitations for sign language reconstruction.
%They rely on general-purpose pose detectors and body priors that are not tailored to the structured and semantically rich articulations of sign languages, leading to domain shift, depth ambiguity, and over-regularization.
They rely on general-purpose pose detectors and body priors not tailored to sign language articulations, leading to domain shift, depth ambiguity, and over-regularization.
Moreover, these methods are trained on everyday human motion and lack exposure to the linguistic and cultural nuances of sign languages.
\section{Method}
We introduce DexAvatar, a method for reconstructing 3D whole-body pose and mesh from monocular sign language videos.
In sign languages, meaning is conveyed through the coordinated use of hand gestures, facial expressions, and upper body movements within 3D space.
To represent these modalities jointly, we employ the SMPL-X~\cite{smplx} parametric model.
DexAvatar, similar to SMPLify-X~\cite{smplx}, is an optimization-based method that uses pseudo-ground truth from off-the-shelf tools, including initial camera parameters, SMPL-X parameters, 2D keypoints, and 3D hand parameters~\cite{cai2023smpler, khirodkar2024sapiens, pavlakos2024reconstructing}.
%Since off-the-shelf tools that are not dedicated to sign language tasks may fail to capture the fine-grained hand articulations and upper-body expressions in signing, we introduce specialized priors, SignHposer for hands and SignBPoser for body.
Since off-the-shelf tools are not dedicated to sign language and may fail to capture fine-grained hand articulations and upper-body expressions, we introduce specialized priors: SignHPoser for hands and SignBPoser for body.

\begin{figure*}[t]
\centering
    \includegraphics[width=1\textwidth]{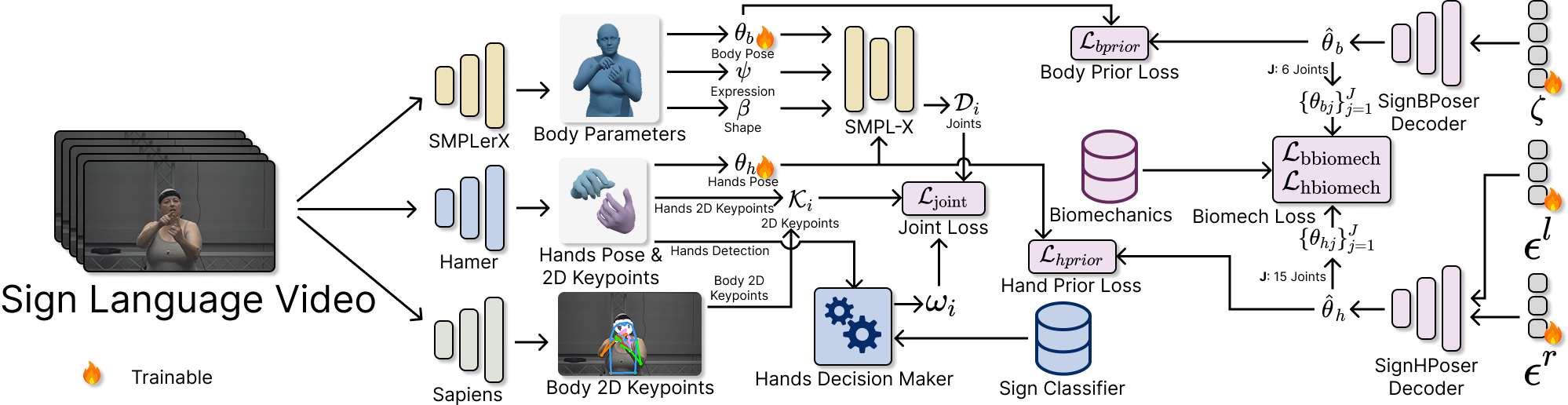}
    \caption{\textbf{Overview of the DexAvatar pipeline}. Given a set of input frames, we first run SMPLerX~\cite{cai2023smpler} and HaMeR~\cite{pavlakos2024reconstructing} to obtain initial body and hand pose estimates.
    We then refine these estimates by fitting to a 2D joint, using Sapiens~\cite{khirodkar2024sapiens} for body keypoints and HaMeR for hand keypoints by minimizing the reprojection error ($\mathcal{L}_{\text {joint }}$) to the detected joints $\mathcal{K}_i$.
    To generate plausible hand and body articulations, we constrain poses to learned manifolds, where SignBPoser maps a body latent $\zeta$ to $\theta_b$, and SignHPoser maps independent left and right latents $\epsilon^{\ell}$ and $\epsilon^{r}$ to $\theta_h$.
    Finally, bio-mechanical constraints enforce physically plausible articulation, producing accurate 3D signing avatars.
    } \label{fig:arch}
\end{figure*}

\subsection{Preliminaries}

%\noindent\textbf{SMPLify-X} is a widely used optimization method based on SMPL-X that reconstructs 3D body and hand poses, and facial expression from monocular images (see section 1 in Supplementary material for a comprehensive discussion on SMPL-X).
\noindent\textbf{SMPLify-X} is a widely used optimization method based on SMPL-X that reconstructs 3D body and hand poses and facial expressions from monocular images (see Section 1 in the Supplementary Material).
The reconstruction is achieved by minimizing the following objective function with respect to $\boldsymbol{\beta}$, $\boldsymbol{\psi}$, and $\boldsymbol{\theta}$:

\begin{equation}
\label{eq:smplifyx_loss}
\mathcal{L}
=
\mathcal{L}_{\text{joint}}
+ \lambda_{\zeta}\,\mathcal{L}_{\zeta}
+ \lambda_{\text{pen}}\,\mathcal{L}_{\text{pen}}.
\end{equation}

$\mathcal{L}_{\text{joint}}$, defined in Eq.~\eqref{eq:2djoint}, minimizes the error between the detected 2D keypoints $\mathcal{K}$ and the corresponding 3D keypoints $P(\mathcal{D})$ projected onto the image plane via $P(\cdot)$.
$\mathcal{J}$ represents the set of 3D joints, and $P(\cdot)$ projects a 3D joint $\mathcal{D}_i \in \mathbb{R}^3$ from the world coordinate system to the 2D image coordinates.
The term $\mathcal{K}_i$ represents the corresponding 2D keypoint, which may come from pseudo or ground truth annotations.
$\omega_i$ denotes the confidence of $\mathcal{K}_i$, $\gamma_i$ is a predefined weight for joint $\mathcal{D}_i$, and $\psi$ represents a robust Geman-McClure loss function~\cite{fessler2000statistical} to prevent the disturbance from noisy supervision signals. The objective is given as:

\begin{equation}
    \label{eq:2djoint}
    \mathcal{L}_{\text {joint }}=\frac{1}{|\mathcal{J}|} \sum_{i \in \mathcal{J}} \gamma_i \omega_i \psi\left(P\left(\mathcal{D}_i\right)-\mathcal{K}_i\right).
\end{equation}

$\mathcal{L}_{\zeta}$, defined in Eq.~\eqref{eq:vposerembed}, is a zero-mean Gaussian prior on the VPoser learnable body pose embedding $\zeta \in \mathbb{R}^{d}$, where the negative log-prior yields an $\ell_2$ penalty on each latent dimension.
Here, $i\in\{1,\dots,d\}$ indexes the latent dimensions and $\sigma_i^{2}$ denotes the variance of the $i$-th dimension under the diagonal Gaussian prior. The loss term is added to the total objective with weight $\lambda_{\zeta}$.

\begin{equation}
    \label{eq:vposerembed}
    \mathcal{L}_{\zeta}
    = \sum_{i=1}^{d} \frac{\zeta_i^{2}}{\sigma_i^{2}},
\qquad \text{with } \sigma_i^{2} = 1 \ \forall\, i.
\end{equation}

$\mathcal{L}_{\text{pen}}$, defined in Eq.~\eqref{eq:prelim-pen}, prevents self-collision by first detecting colliding face pairs $\mathcal{C}$ with a bounding volume hierarchy and then penalizing the bi-directional intrusion depth for each pair $(f_s,f_t)$.
For any face $f$, $\mathcal{V}(f)$ is its vertex set and $\Psi_f(v)$ is the conic signed distance to $f$ (negative inside), so $\max(0,-\Psi_f(v))$ measures how far a vertex penetrates.
The loss sums these depths for vertices of $f_s$ in the $f_t$ field and vice versa, optionally normalizing by $|\mathcal{C}|$, and is added to the total objective with weight $\lambda_{\text{pen}}$.

\begin{multline}
    \label{eq:prelim-pen}
    \mathcal{L}_{\text{pen}}
    = \frac{1}{|\mathcal{C}|}
    \sum_{(f_s,f_t)\in\mathcal{C}}
    \Big[\sum_{v\in \mathcal{V}(f_s)} (\max(0,-\Psi_{f_t}(v)))^2 \\
    + \sum_{v\in \mathcal{V}(f_t)} (\max(0,-\Psi_{f_s}(v)))^2
    \Big].
\end{multline}

\subsection{Data Preprocessing}
\subsubsection{Body Data}
\label{subsec:Data_body}
\begin{figure}[t]
\centering
    \includegraphics[width=.4\textwidth]{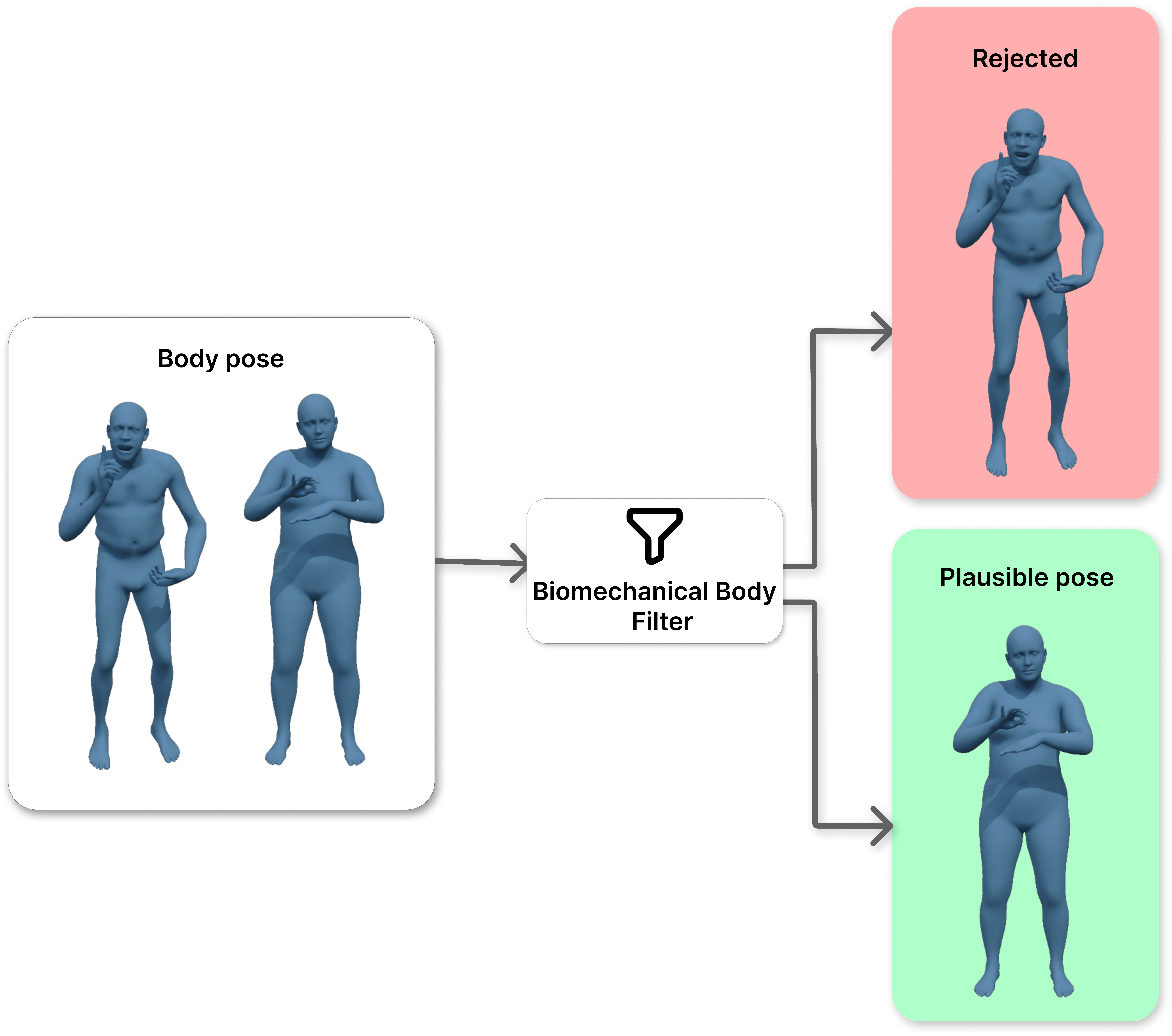}
\caption{\textbf{Bio-mechanical body filter}. For body data from ~\cite{how2sign}, we enforce joint range of motion and signer space constraints on shoulders, elbows/forearm, and wrists. Frames that violate these envelopes are rejected, and only plausible body poses are retained for training.}
\label{fig:bprior_filtering}
\end{figure}

%We used the 3D body data provided by SignAvatars~\cite{yu2024signavatars}, derived from the How2Sign~\cite{how2sign} dataset.
We use the 3D body data provided by SignAvatars~\cite{yu2024signavatars}, derived from the How2Sign~\cite{how2sign} dataset.
%As these data constitute pseudo-ground truth, they may contain residual noise and bias in the recovered motions.
Since these data constitute pseudo-ground truth, they may contain residual noise and bias in the recovered motions.
%Training the body pose prior, SignBPoser, on such data without additional safeguards can degrade performance.
Training SignBPoser on such data without additional safeguards can degrade performance.
%Therefore, our approach begins with filtering out implausible poses using established theory for bio-mechanical constraints of joints in the human body~\cite{hamill2006biomechanical,knudson2007fundamentals} as shown in Fig.~\ref{fig:bprior_filtering}.
Therefore, we filter out implausible poses using established bio-mechanical constraints of human body joints~\cite{hamill2006biomechanical,knudson2007fundamentals}, as shown in Fig.~\ref{fig:bprior_filtering}.
Since major movements in sign languages employ a subset of body joints, that is, shoulders, elbows/forearms, and wrists, we focus on those in our preprocessing.
%For those joints, we define plausible ranges of motion, constrained by physiological degrees of freedom and the theory of signer space~\cite{branchini2020grammar,wilcox2020conceptualization}, a torso-centric 3D region where signs are produced and perceived (see section 2 of Supplementary material for details).
For these joints, we define plausible ranges of motion constrained by physiological degrees of freedom and signer space~\cite{branchini2020grammar,wilcox2020conceptualization}—a torso-centric 3D region where signs are produced and perceived (see Section 2 of Supplementary Material for details).
%Specifically, we remove frames from the 3D body data where joint angles are outside the defined ranges of motion, such as overly elevated, excessively retracted, or fully outstretched arms, inconsistent with real-life signing.
Specifically, we remove frames where joint angles fall outside the defined ranges of motion, such as overly elevated, excessively retracted, or fully outstretched arms that are inconsistent with real-life signing.

\subsubsection{Hand Data}
\label{subsec:Data_hand}
\noindent\textbf{Data Acquisition.}
%We recorded sign language motion capture data to train the hand pose prior SignHPoser.
We recorded sign language motion capture data to train SignHPoser, our hand pose prior.
%We used a Vicon setup~\cite{vicon_devices} that consists of 9 high-resolution cameras, strategically placed to ensure full coverage of the signer's movements.
We used a Vicon setup~\cite{vicon_devices} consisting of 9 high-resolution cameras strategically placed to ensure full coverage of the signer's movements.
Furthermore, we used Manus gloves~\cite{manus_metagloves_pro} to track finger articulations.
%We collected fingerspelling data from 8 signers, six proficient in Australian Sign Language (Auslan) and two fluent in American Sign Language (ASL).
We collected fingerspelling data from 8 signers: six proficient in Australian Sign Language (Auslan) and two fluent in American Sign Language (ASL).
Each participant spelled a curated list of 93 words letter-by-letter.
%The raw motion capture data were retargeted onto an SMPL-X rig in Blender using the Rokoko plugin.
The raw motion capture data were retargeted to an SMPL-X rig in Blender using the Rokoko plugin.
%To preserve global hand motion and enable realistic upper body movements, inverse kinematics constraints were introduced, allowing the arms to follow the wrist trajectories driven by the glove data.
To preserve global hand motion and enable realistic upper body movements, we introduced inverse kinematics constraints that allow the arms to follow the wrist trajectories driven by the glove data.
%The resulting animations were baked into an SMPL-X armature (Details of the retargeting are provided in section 3 of the Supplementary material).
The resulting animations were baked into an SMPL-X armature (details of the retargeting are provided in Section 3 of the Supplementary Material).

\begin{figure}[t]
\centering
    \includegraphics[width=.45\textwidth]{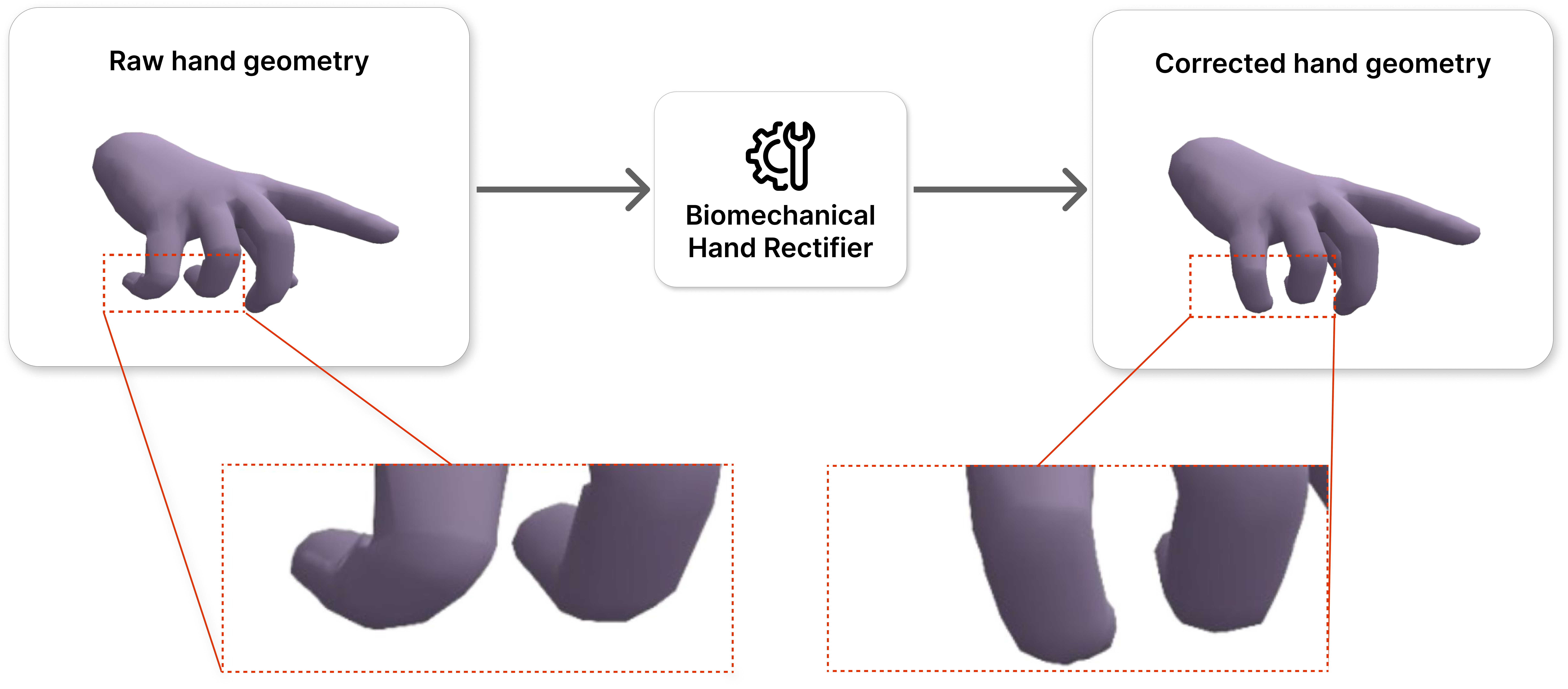}
\caption{\textbf{Bio-mechanical hand rectifier}. For raw mocap hand data, we correct implausible joint configurations by enforcing per-joint limits on bending, splaying, and twisting (15 hand joints). The rectifier outputs corrected hand geometry for training.}
\label{fig:hprior_filtering}
\end{figure}

\noindent\textbf{Hand Data Preprocessing.}
%Motion capture data are prone to noise due to errors in sensor readings and tracking, resulting in implausible hand poses.
Motion capture data are prone to noise from sensor reading and tracking errors, resulting in implausible hand poses.
%To address such issues, we corrected the dataset by using bio-mechanical constraints for the hands, as shown in Fig.~\ref{fig:hprior_filtering}.
To address this, we correct the dataset using bio-mechanical constraints for the hands, as shown in Fig.~\ref{fig:hprior_filtering}.
%We leveraged the established theory of hand bio-mechanics~\cite{chen2013constraint} and defined constraints for each hand joint.
We leverage established hand bio-mechanics theory~\cite{chen2013constraint} and define constraints for each hand joint.
%Joint constraints can be described with 3 Euler angles corresponding to joint bending, splaying, and twisting.
Joint constraints are described using 3 Euler angles corresponding to joint bending, splaying, and twisting.
%Since the joint rotation coordinates used by MANO differ from the theoretical constraints, we aligned the axis following previous work~\cite{zhang2024weakly}.
Since the joint rotation coordinates used by MANO differ from the theoretical constraints, we align the axes following previous work~\cite{zhang2024weakly}.

\subsection{Body and Hand Pose Priors}
%Vposer~\cite{smplx} is a body prior trained on three publicly available human motion capture datasets, CMU~\cite{cmumocap}, the train set of Human3.6M~\cite{ionescu2013human3}, and the PosePrior dataset~\cite{akhter2015pose}.
VPoser~\cite{smplx} is a body prior trained on three publicly available human motion capture datasets: CMU~\cite{cmumocap}, the training set of Human3.6M~\cite{ionescu2013human3}, and the PosePrior dataset~\cite{akhter2015pose}.
%Vposer is used as a prior in previous work to restrict the body pose parameters within an acceptable range of human motion.
VPoser is used as a prior in previous work to constrain body pose parameters to an acceptable range of human motion.
%Vposer has shown exceptional performance in reconstructing a whole-body mesh from in-the-wild videos.
VPoser has shown strong performance in reconstructing whole-body meshes from in-the-wild videos.
%We argue that using such general-purpose priors in the specific case of signing can generate motions falling outside the signer space~\cite{branchini2020grammar,wilcox2020conceptualization}.
However, we argue that using such general-purpose priors for signing can generate motions that fall outside signer space~\cite{branchini2020grammar,wilcox2020conceptualization}.
%Therefore, we train SignBPoser and SignHPoser using our pre-processed sign language datasets.
Therefore, we train SignBPoser and SignHPoser on our preprocessed sign language datasets.
%Both priors use variational autoencoder (VAE)~\cite{kingma2013auto} architectures.
Both priors use a variational autoencoder (VAE)~\cite{kingma2013auto} architecture.
%This enables learning of compact latent representations of sign language hand and body poses by enforcing regularization toward an isotropic Gaussian distribution in the latent space, supporting realistic pose synthesis and efficient optimization in downstream tasks.
This enables learning compact latent representations of sign language hand and body poses by enforcing regularization toward an isotropic Gaussian distribution in the latent space, which supports realistic pose synthesis and efficient optimization in downstream tasks.
The training loss is:

\begin{equation}
\label{eq:total}
\begin{aligned}
\mathcal{L} &= c_1 \mathcal{L}_{\text{KL}} + c_2 \mathcal{L}_{\text{recon}} + c_3 \mathcal{L}_{\text{mesh}} + c_4 \mathcal{L}_{\text{orth}} \\
&\quad + c_5 \mathcal{L}_{\text{reg}} + c_6 \mathcal{L}_{\text{biomech}},
\end{aligned}
\end{equation}
where $c_1$, $c_2$, $c_3$, $c_4$, $c_5$, and $c_6$ are loss weights which are empirically set to 0.001, 0.999, 0.999, 0.01, 0.0001, and 1.5 for SignBPoser and 0.0001, 0.999, 0.999, 0.01, 0.0001, and 1.5 for SignHPoser, respectively.

$\mathcal{L}_{\text{KL}}$, defined in Eq.~\eqref{eq:kl}, regularizes the latent code $\boldsymbol{Z}\in\mathbb{R}^{33}$ toward a standard normal distribution.
The encoder posterior is $q(\boldsymbol{Z}\mid \boldsymbol{R})$ where $R\in\mathbb{R}^{3\times3}$ is the input rotation matrix for each joint and $\mathcal{N}(0,I)$ denotes the standard normal distribution,

\begin{equation}
\label{eq:kl}
\mathcal{L}_{\text{KL}} = \mathrm{KL}\!\left(q(\boldsymbol{Z}\mid\boldsymbol{R}) \,\|\, \mathcal{N}(0,I)\right).
\end{equation}

$\mathcal{L}_{\text{recon}}$, defined in Eq.~\eqref{eq:recon}, measures the squared $\ell_2$ error between the input axis–angle vector $\boldsymbol{\alpha}$ and its reconstruction $\hat{\boldsymbol{\alpha}}$,

\begin{equation}
\label{eq:recon}
\mathcal{L}_{\text{recon}} = \| \boldsymbol{\alpha} - \hat{\boldsymbol{\alpha}} \|_2^2.
\end{equation}

$\mathcal{L}_{\text{mesh}}$, defined in Eq.~\eqref{eq:mesh}, enforces vertex-level fidelity between the predicted mesh $\hat{M}$ from the SMPL-X layer and the reference mesh $M$, measured as a per-vertex squared $\ell_2$ error,

\begin{equation}
\label{eq:mesh}
\mathcal{L}_{\text{mesh}} = \| M - \hat{M} \|_2^2.
\end{equation}

$\mathcal{L}_{\text{orth}}$, defined in Eq.~\eqref{eq:orth}, constrains rotations to be valid by enforcing orthogonality and unit determinant, where $\hat{R}\in\mathbb{R}^{3\times3}$ denotes the output rotation matrix for each joint and $I$ is the $3\times3$ identity,

\begin{equation}
\label{eq:orth}
\mathcal{L}_{\text{orth}} = \| \hat{R}\hat{R}^{\top} - I \|_2^2.
\end{equation}

$\mathcal{L}_{\text{reg}}$, defined in Eq.~\eqref{eq:reg}, discourages overfitting by penalizing the $\ell_2$ norm of the trainable parameters $\boldsymbol{\phi}$,

\begin{equation}
\label{eq:reg}
\mathcal{L}_{\text{reg}} = \| \boldsymbol{\phi} \|_2^2.
\end{equation}

$\mathcal{L}_{\text{biomech}}$, defined in Eq.~\eqref{eq:biomech}, enforces per-joint anatomical limits.
For joint $j$, let $\overline{\boldsymbol{\theta}}_{j,\min}$ and $\overline{\boldsymbol{\theta}}_{j,\max}$ denote the lower and upper angular bounds.
The penalty is zero when $\boldsymbol{\theta}_j$ lies within $[\overline{\boldsymbol{\theta}}_{j,\min},\,\overline{\boldsymbol{\theta}}_{j,\max}]$ and grows quadratically outside this range. $\max$ is taken element-wise.
We apply this to $J=6$ body joints for SignBPoser and $J=15$ hand joints for SignHPoser,

\begin{equation}
\label{eq:biomech}
\mathcal{L}_{\text{biomech}}
= \sum_{j=1}^{J}
\big\|
\max\!\big\{
\boldsymbol{\theta}_j - \overline{\boldsymbol{\theta}}_{j,\max},\,
\overline{\boldsymbol{\theta}}_{j,\min} - \boldsymbol{\theta}_j,\,
\mathbf{0}
\big\}
\big\|_2^2.
\end{equation}

\subsection{Optimization}
Initializing with reliable SMPL-X estimates improves avatar reconstruction from videos.
%We therefore seed the DexAvatar optimization with off-the-shelf SMPL-X parameters, 2D keypoints, and camera estimates~\cite{khirodkar2024sapiens,cai2023smpler,yang2023effective} as shown in Fig.~\ref{fig:arch}.
We therefore initialize DexAvatar optimization with off-the-shelf SMPL-X parameters, 2D keypoints, and camera estimates~\cite{khirodkar2024sapiens,cai2023smpler,yang2023effective}, as shown in Fig.~\ref{fig:arch}.
%These detectors often struggle with sign language data, requiring additional improvements.
However, these detectors often struggle with sign language data, requiring additional improvements.
%Hence, the optimization objective of DexAvatar follows Eq.~\eqref{eq:smplifyx_loss} by retaining the joint loss and the interpenetration term from SMPLify-X, while replacing the generic VPoser prior with SignBPoser for the body and adding SignHPoser prior for the hands.
Hence, the DexAvatar optimization objective follows Eq.~\eqref{eq:smplifyx_loss} by retaining the joint loss and interpenetration term from SMPLify-X, while replacing the generic VPoser prior with SignBPoser for the body and adding the SignHPoser prior for the hands.
%In addition, we introduce temporal consistency and bio-mechanical penalties for hand and body, tailored to signing dynamics.
In addition, we introduce temporal consistency and bio-mechanical penalties for the hand and body, tailored to signing dynamics.
%We restrict lower-body joints and deactivate the non-dominant arm for one-handed signs to improve stability.
We also restrict lower-body joints and deactivate the non-dominant arm for one-handed signs to improve stability.
The optimization minimizes the following objective:

\begin{equation}
\label{eq:fitting_equation}
\begin{split}
\mathcal{L} = \mathcal{L}_{\text{joint}} + \lambda_1 \mathcal{L}_{\text{bprior}} + \lambda_2 \mathcal{L}_{\text{hprior}}
+ \lambda_3 \mathcal{L}_{\text{pen}} \\ 
+ \lambda_4 \mathcal{L}_{\text{temp}}
+ \lambda_5 \mathcal{L}_{\text{bbiomech}}
+ \lambda_6 \mathcal{L}_{\text{hbiomech}}.
\end{split}
\end{equation}

$\mathcal{L}_{\text{joint}}$ adopts the formulation in Eq.~\eqref{eq:2djoint}, leveraging 2D keypoints from Sapiens~\cite{khirodkar2024sapiens} for the body and HaMeR~\cite{pavlakos2024reconstructing} for the hands.
Since signing typically involves minimal to no lower-body motion, we set $\omega_i = 0$  in Eq.~\eqref{eq:2djoint} for all lower body joints to exclude them from optimization.
Furthermore, we utilize the sign classifier from~\cite{forte2023reconstructing} to distinguish between one-handed and two-handed signs.
In the case of one-handed signing, our Hand Decision Maker (see Fig.~\ref{fig:arch}) disables optimization of the non-dominant arm (shoulder, elbow, and wrist) and non-dominant hand by assigning $\omega_i = 0$ in Eq.~\eqref{eq:2djoint}.
This strategy prevents spurious updates and ensures the optimization focuses only on the active parts of the body where refinement is necessary.

$\mathcal{L}_{\text{bprior}}$ (for the body), defined in Eq.~\eqref{eq:bp_loss}, integrates SignBPoser to regularize infeasible body poses.
Since SignBPoser is trained on sign language data, it provides compact mapping from low-dimensional $\bar\zeta$ to axis-angle representation of the body pose $\theta_b$.
To better optimize the low-dimensional embedding, the estimated body pose $\hat{\theta_b}$ from SMPLerX~\cite{cai2023smpler} is treated as a supervisory signal, in coordination with the added regularization term $\bar\zeta$ similar to Eq.~\eqref{eq:vposerembed} with a weighting of $\lambda_{\bar\zeta}$.
The body prior loss term is formulated as follows,

\begin{equation}
    \label{eq:bp_loss}
    \mathcal{L}_{bprior} = \psi\left(\theta_b - \hat{\theta}_b\right) + \lambda_{\bar\zeta}\,\mathcal{L}_{\bar\zeta}.
\end{equation}

$\mathcal{L}_{\text{hprior}}$ (for the hand), defined in Eq.~\eqref{eq:bh_loss}, integrates SignHPoser to regularize infeasible hand poses.
We follow the same formulation as Eq.~\eqref{eq:bp_loss}.
We treat the left and right hand pose $\hat{\theta_h}$ from HaMeR~\cite{pavlakos2024reconstructing} as a supervisory signal along with $\epsilon$, which is the low-dimensional representation for each hand (see Eq.~\eqref{eq:vposerembed}).

\begin{equation}
    \label{eq:bh_loss}
    \mathcal{L}_{hprior} = \psi\left(\theta_h - \hat{\theta}_h\right) + \lambda_{\epsilon^{l}}\,\mathcal{L}_{\epsilon^{l}} + \lambda_{\epsilon^{r}}\,\mathcal{L}_{\epsilon^{r}}.
\end{equation}

$\mathcal{L}_{\text{temp}}$ ensures temporal consistency across frames, and incorporates the pose parameters from the previous frame, denoted as $\boldsymbol{\theta}^{\text{pre}}_b$.
We enforce smooth transitions between frames by penalizing discrepancies in the motion of corresponding joints.

\begin{equation}
\label{eq:temporal_loss}
    \mathcal{L}_{\text{temp}} = \psi\left(\theta_b - {\theta}_b^{\text{pre}}\right).
\end{equation}

$\mathcal{L}_{\text{bbiomech}}$ and $\mathcal{L}_{\text {hbiomech}}$ incorporate bio-mechanical loss constraints for the body~\cite{branchini2020grammar,wilcox2020conceptualization} and the hands~\cite{chen2013constraint} as additional supervision during optimization.
The associated loss term is unchanged and follows Eq.~\eqref{eq:biomech}.
\section{Experiments}
\noindent\textbf{Datasets.} We evaluated DexAvatar on the motion capture dataset used in SGNify~\cite{forte2023reconstructing}.
The dataset consists of 57 German signs.
Following the standard evaluation protocol for this dataset, we evaluate on central portions of each sign from the raw videos and compute the quantitative results on only these central frames (in total, 2,872 frames).

\noindent\textbf{Evaluation Metrics.} Following prior work SGNify~\cite{forte2023reconstructing} and Neural Sign Actors~\cite{baltatzis2024neural}, we evaluated DexAvatar using mean vertex-to-vertex error (TR-V2V), restricting computation to vertices above the pelvis.
We report results for three regions: Upper Body (excluding the face), Left Hand, and Right Hand.
To assess SignBPoser and SignHPoser independently, we compute MPJPE and MPVPE on the recovered joints and meshes.
% For analyzing the effectiveness of SignBPoser within the optimization, we evaluate over four vertex subsets: Full Body (10,475 vertices), Upper Body (vertices above the pelvis), Upper Body (-H) (above-pelvis without the head), and Upper Body (-F) (above-pelvis without the face).

\noindent\textbf{Implementation Details.} DexAvatar is implemented in Pytorch~\cite{imambi2021pytorch} and optimized using LBFGS~\cite{moritz2016linearly}.
All experiments are performed on NVIDIA-RTX 4090 with 24 GB GPU memory and 64 GB CPU memory.
SignBPoser and SignHPoser are built using an encoder-decoder VAE, each with 3 linear layers with an embedding size of 512.
During training, we use the Adam optimizer~\cite{kinga2015method} and the learning rate of $1e{-3}$.

\begin{table}[t]
    \setlength{\tabcolsep}{1pt}
    \centering
    \caption{\textbf{Quantitative results}. We compare DexAvatar with the current state-of-the-art methods on TR-V2V error (mm). We report results in three regions, \ie, UBody(-F): Upper Body excluding the face, LHand: Left Hand, and RHand: Right Hand. EVA* denotes our modification of EVA~\cite{hu2024expressive} to accommodate one-handed signs.}
    \label{tab:main_results}
    \begin{tabular}{lccc}
        \toprule
        \textbf{Method} & \textbf{UBody (-F)}$\downarrow$ & \textbf{LHand}$\downarrow$ & \textbf{RHand}$\downarrow$ \\
        \midrule
        FrankMoCap~\cite{rong2021frankmocap} & 78.07  & 20.47     & 19.62   \\
        PIXIE~\cite{feng2021collaborative} &  60.11 & 25.02     & 22.42   \\
        PyMAF-X~\cite{zhang2023pymaf} & 68.61  &  21.46    & 19.19   \\
        SMPLify-SL~\cite{forte2023reconstructing} &  56.07  & 22.23     & 18.83   \\
        SGNify~\cite{forte2023reconstructing} &  55.63  &  19.22   & 17.50  \\
        OSX~\cite{lin2023one} & 47.32  &   18.34  &  18.12 \\
        Neural Sign Actors~\cite{baltatzis2024neural} & 46.42 & 16.17  & 15.23\\
        EVA* & 40.38 & 13.73 & 13.68 \\
        \midrule
        \textbf{DexAvatar (Ours)} & \textbf{30.13} & \textbf{13.53} & \textbf{13.08} \\
        \bottomrule
    \end{tabular}
\end{table}

\section{Results}
\subsection{Quantitative Results}
Table~\ref{tab:main_results} reports a comparison of DexAvatar against state-of-the-art baselines on the SGNify dataset.
DexAvatar achieves the best performance across regions, outperforming FrankMocap~\cite{rong2021frankmocap}, PyMAF-X~\cite{zhang2023pymaf}, PIXIE~\cite{feng2021collaborative}, SMPLify-SL~\cite{forte2023reconstructing}, SGNify~\cite{forte2023reconstructing}, OSX~\cite{lin2023one}, and Neural Sign Actors~\cite{baltatzis2024neural}.
As the dataset contains both one- and two-handed signing, we modify EVA~\cite{hu2024expressive} to accommodate one-handed signs, and denote it as EVA*.
DexAvatar surpasses Neural Sign Actors on the left and right hands by 16.32\% and 14.11\%, respectively, and yields a substantial \textbf{35.11\%} improvement on the upper body.

\begin{table}[t]
    \setlength{\tabcolsep}{2pt}
    \centering
    % \scriptsize
    \caption{We conduct an ablation study to understand the effectiveness of SignBPoser within DexAvatar. We conduct experiments with three different variants. $\text{BP}_{\text{u}}$: trained on unfiltered data, $\text{BP}_{\text{f}}$: trained on bio-mechanically filtered data, $\text{BP}_{\text{f+bio}}$: trained on filtered data and with body bio-mechanical loss. We evaluate over four vertex subsets: FBody: Full Body (10,475 vertices), UBody: Upper Body (vertices above the pelvis), UBody (-H): Upper Body without head (above-pelvis without the head).}
    \label{tab:ablation_fit_body}
    \begin{tabular}{lcccc}
        \toprule
        \textbf{Method} & \textbf{FBody}$\downarrow$ & \textbf{UBody}$\downarrow$ & \textbf{UBody (-H)}$\downarrow$ & \textbf{UBody (-F)}$\downarrow$ \\
        \midrule
        $\text{BP}_{\text{u}}$ & 43.18 & 29.95 & 44.72 & 34.06 \\
        $\text{BP}_{\text{f}}$ & \textbf{42.32} & \textbf{26.78} & \textbf{41.35} & \textbf{30.28} \\
        $\text{BP}_{\text{f+bio}}$ & 42.38 & 26.93 & 41.88 & 30.44 \\
        % $\text{BP}_{\text{f}}$ + $\text{Dex}_{\text{b}}$ & \textbf{42.25} & \textbf{26.68} & \textbf{41.33} & \textbf{30.18} \\    
        % $\text{BP}_{\text{fb}}$ + $\text{Dex}_{\text{b}}$ & 42.58 & 27.04 & 41.96 & 30.51 \\
        \bottomrule
    \end{tabular}
\end{table}

% \subsection{Ablation Study}

% In this subsection, we present ablation results to verify the effectiveness of our SignBPoser and SignHPoser.
% All evaluations are on the SGNify dataset.

\noindent\textbf{Effectiveness of SignBPoser.}
Table~\ref{tab:ablation_fit_body} summarizes ablation results of the body prior, SignBPoser, within DexAvatar.
%We perform a hyperparameter search (see section 4 of Supplementary Material  for details) for three variants of the body prior and report results based on the best configurations for $\text{BP}_{\text{u}}$ trained on unfiltered data, $\text{BP}_{\text{f}}$ trained on bio-mechanically filtered data, and $\text{BP}_{\text{f+bio}}$ trained on filtered data and with body bio-mechanical loss.
We perform a hyperparameter search (see section 4 of Supplementary Material for details) for three variants and report results based on the best configurations: $\text{BP}_{\text{u}}$ trained on unfiltered data, $\text{BP}_{\text{f}}$ trained on bio-mechanically filtered data, and $\text{BP}_{\text{f+bio}}$ trained on filtered data with bio-mechanical loss.
% The selected configuration is identical across variants, with KL weight $10^{-3}$, width $512$, and latent dimensionality $33$.
% The bio-mechanical loss is enabled with a weight of 1.5 only for $\text{BP}_{\text{f+bio}}$, and omitted for $\text{BP}_{\text{u}}$ and $\text{BP}_{\text{f}}$.
% These settings are used in all downstream experiments.
The first two rows of Table~\ref{tab:ablation_fit_body} show the results for $\text{BP}_{\text{u}}$ and $\text{BP}_{\text{f}}$.
$\text{BP}_{\text{f}}$ consistently outperforms $\text{BP}_{\text{u}}$ across all metrics, with relative error reductions of \textbf{2.0\%} (FBody), \textbf{10.6\%} (UBody), \textbf{7.5\%} (UBody (–H)), and \textbf{11.1\%} (UBody (–F)).
This establishes the efficacy of our data preprocessing using bio-mechanical constraints.
Using a prior trained on the filtered data with bio-mechanical loss
in $\text{BP}_{\text{f+bio}}$ yields a slight degradation (\(+0.14\%\) FBody, \(+0.56\%\) UBody, \(+1.28\%\) UBody (–H), \(+0.53\%\) UBody (–F), suggesting mild over-regularization during optimization.
Finally, adding bio-mechanical loss during optimization (excluded from Table~\ref{tab:ablation_fit_body}) while retaining $\text{BP}_{\text{f}}$ produces the best results on all subsets, with relative error reductions of \textbf{0.17\%} (FBody), \textbf{0.37\%} (UBody), \textbf{0.05\%} (UBody (–H)), and \textbf{0.33\%} (UBody (–F)) compared to $\text{BP}_{\text{f}}$.

% In contrast, adding a lightweight 

% Finally, replacing  Sbp$_{fb}$ with Sbp$_{f}$ and retaining the biomechanical loss offers no additional gain.
% We therefore adopt ${\text{b}}_{\text{sbp\_f\_bml}}$ for all subsequent experiments.

\begin{table}[t]
    \setlength{\tabcolsep}{2pt}
    \centering
    \caption{Ablation study of SignHPoser within \mymethod.}
    \label{tab:signhposer_variants}
    \begin{tabular}{lccc}
        \toprule
        \textbf{Method} & \textbf{UBody (-F)}$\downarrow$ & \textbf{LHand}$\downarrow$ & \textbf{RHand}$\downarrow$ \\
        \midrule
        $\text{HP}_{\text{u}}$ & 31.34 & 14.19 & 13.92 \\
        $\text{HP}_{\text{f}}$ & 30.17 & 13.55 & 13.06 \\
        $\text{HP}_{\text{f+bio}}$ & \textbf{30.13} & \textbf{13.53} & \textbf{13.08} \\
        % ${\text{bh}}_{\text{shp\_f\_bml}}$ & 30.30 & 13.57 & 13.15 \\
        % ${bh}_{full}$ & 30.27 & 13.57 & 13.34 \\
        \bottomrule
    \end{tabular}
\end{table}

\noindent\textbf{Effectiveness of SignHPoser.}
We evaluate the hand prior SignHPoser using the best performing configuration for the body prior, that is $\text{BP}_{\text{f}}$ and body-based bio-mechanical loss during optimization.
Table~\ref{tab:signhposer_variants} summarizes ablation results.
We perform a hyperparameter search (see section 4 of Supplementary Material  for details) for three hand prior variants and report results based on the best configurations for $\text{HP}_{\text{u}}$ trained on uncorrected hand data, $\text{HP}_{\text{f}}$ trained on the bio-mechanically corrected data, and $\text{HP}_{\text{f+bio}}$ trained on the corrected data with hand bio-mechanical loss.
% The selected configurations are $\text{HP}_{\text{u}}$ with KL weight $10^{-4}$, width $512$, latent $24$, $\text{HP}_{\text{f}}$ with KL weight $10^{-4}$, width $512$, latent $23$, and $\text{HP}_{\text{f+bio}}$ with KL weight $10^{-4}$, width $512$, latent $23$ with the biomechanical loss enabled with a weight of 1.5.
% Note, we retain the body prior ${\text{b}}_{\text{sbp\_f\_bml}}$ from the previous experiment to assess the combined effect on performance.
The first two rows of Table~\ref{tab:signhposer_variants} show the results for $\text{HP}_{\text{u}}$ and $\text{HP}_{\text{f}}$ variants.
It can be observed that $\text{HP}_{\text{f}}$ outperforms $\text{HP}_{\text{u}}$ on all metrics, with relative error reductions of \textbf{3.7\%} on Upper Body, \textbf{4.5\%} on Left Hand, and \textbf{6.2\%} on Right Hand.
This demonstrates the importance of the correction process using bio-mechanical constraints on the hand data.
Adding a bio-mechanical regularizer to the filtered prior in $\text{HP}_{\text{f+bio}}$ increases accuracy compared to $\text{HP}_{\text{f}}$, yielding slight improvements on Upper Body (0.13\%) and Left Hand (0.15\%).
However, Right Hand performance degrades slightly (0.2\%).
This indicates that introducing bio-mechanical constraints provides useful physical regularization in the fitting process. Please see section 5 of Supplementary Material for ablation of SignHPoser with Vposer. 
% For the last two rows, adding a biomechanical loss during fitting ${\text{bh}}_{\text{shp\_f,bml}}$ makes the errors across all the metrics go slightly higher.
% Adding Shp$_{fhb}$ along with biomechanical loss in ${\text{bh}}_{\text{full}}$ also does not help and even raises the right-hand error.

\subsection{Qualitative Results}
\begin{figure*}[t]
   \caption{\textbf{Qualitative results}. We compare 3D holistic human mesh reconstruction methods on SGNify~\cite{forte2023reconstructing} evaluation dataset. DexAvatar produces significantly better SL reconstructions with plausible body and hand poses.}
 \label{fig:main_res}     
\includegraphics[width=\textwidth,height=1.2\textwidth]{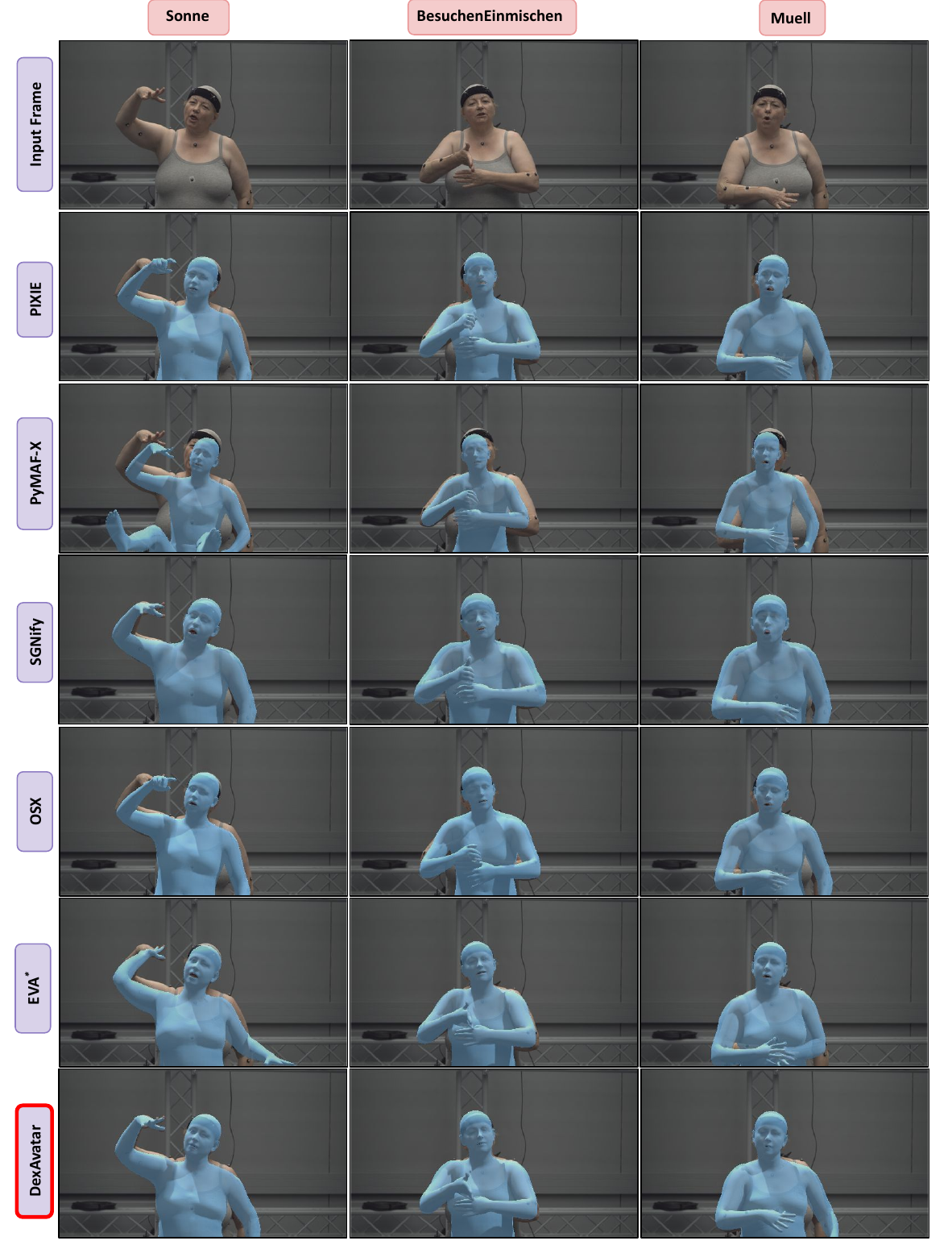}
\end{figure*}

In Fig.~\ref{fig:main_res}, we present qualitative results of \mymethod{} on the SGNify~\cite{forte2023reconstructing} motion capture dataset, compared against existing baselines.
Our approach consistently reconstructs bio-mechanically accurate 3D sign language avatars from monocular videos.
\mymethod{} preserves fine-grained hand articulations and finger orientations that are often missing or distorted in baseline reconstructions.
For the sign \texttt{Sonne}, methods such as PIXIE~\cite{feng2021collaborative}, OSX~\cite{lin2023one}, and EVA* generate avatars that deviate noticeably from ground truth, whereas \mymethod{} maintains close alignment.
%The sign \texttt{BesuchenEinmischen} highlights a complex sequence involving simultaneous hand–body interactions where competing methods including PIXIE~\cite{feng2021collaborative}, PyMAF-X~\cite{zhang2023pymaf}, SGNify~\cite{forte2023reconstructing}, OSX~\cite{lin2023one}, and EVA*~\cite{hu2024expressive} frequently produce misaligned wrists or unnatural limb orientations, while \mymethod{} preserves structural consistency guided by learned hand and body priors.
The sign \texttt{BesuchenEinmischen} highlights a complex sequence involving simultaneous hand–body interactions. Competing methods including PIXIE~\cite{feng2021collaborative}, PyMAF-X~\cite{zhang2023pymaf}, SGNify~\cite{forte2023reconstructing}, OSX~\cite{lin2023one}, and EVA*~\cite{hu2024expressive} frequently produce misaligned wrists or unnatural limb orientations, while \mymethod{} preserves structural consistency guided by our learned hand and body priors.
%For the sign \texttt{Muell}, PyMAF-X and EVA* perform particularly poorly, and other methods struggle with wrist and finger orientations; in contrast, \mymethod{} most closely approximates the ground truth.
For the sign \texttt{Muell}, PyMAF-X and EVA* perform particularly poorly, and other methods struggle with wrist and finger orientations; in contrast, \mymethod{} most closely approximates ground truth.
Overall, these results demonstrate that \mymethod{} produces stable and coherent reconstructions, effectively capturing both global body pose and subtle finger dynamics.
%This leads to avatars that are anatomically plausible and well-aligned with the video. 
This produces avatars that are anatomically plausible and well-aligned with the video.
%We also inspected the evaluation dataset and found implausible hand poses (see section 5 of Supplementary Material for details).
We also inspected the evaluation dataset and found some implausible hand poses in the ground truth (see section 6 of Supplementary Material for details). Please see section 7 of Supplementary Material for additional evaluations on motion blur, noisy, and self-occlusion cases in input frames. 

\section{Conclusion}
We introduced DexAvatar, a method that generates 3D avatars of humans involved in sign language communication from monocular videos.
%Quantitative and qualitative results show consistent gains of DexAvatar over prior works, driven by priors trained on sign language data and an improved optimization pipeline that produces realistic reconstructions.
Quantitative and qualitative results demonstrate consistent improvements over prior work, driven by our sign language-aware priors and an improved optimization pipeline that produces realistic reconstructions.
%These findings highlight the value of task-tailored priors for sign language.
These findings highlight the importance of domain-specific priors for sign language reconstruction.
%Future work will scale data and further strengthen the priors to cover a broader range of signers and signing styles.
Future work will scale the training data and further strengthen the priors to cover a broader range of signers and signing styles.

\section*{Acknowledgements}
\label{sec:acknowledgements}
This research was partly supported by the Australian Government through the Australian Research Council's Discovery Early Career Researcher Award (project DE230100049).
The views expressed herein are those of the authors and are not necessarily those of the Australian Government or Australian Research Council.
We also acknowledge Monash University and National Computational Infrastructure for providing High Performance Computing infrastructure used in this research.
%\newpage
%{
%    \small
 %   \bibliographystyle{ieeenat_fullname}
%    \bibliography{main}
%}

%\clearpage
% \input{camera_ready/supplementary}
%\input{camera_ready/supplementary_new}

\clearpage

\twocolumn[{
\centering
{\large \bfseries Supplementary Material\par}
\vspace{0.8em}
}]

% reset citation numbering for supplement (optional)
\setcounter{section}{0}
\setcounter{figure}{0}
\setcounter{table}{0}

% content
% WACV 2026 Paper Template
% based on the ICCV 2025 template (https://media.eventhosts.cc/Conferences/ICCV2025/ICCV2025-Author-Kit-Feb.zip) with
% WACV-specific details (e.g., 2 tracks) from the WACV 2025 template (https://www.dropbox.com/scl/fi/su44zgdhrzik26p2xu37k/WACV-2025-Author-Kit-Template.zip?rlkey=5qcfimjhxnmx3wlyk7yhk8wg7&dl=0)

\setcounter{page}{1}
% \maketitlesupplementary

\setcounter{section}{0}
\setcounter{figure}{0}
\setcounter{table}{0}
\renewcommand*{\thesection}{S\arabic{section}}
\renewcommand*{\thefigure}{S\arabic{figure}}
\renewcommand*{\thetable}{S\arabic{table}}
%\externaldocument[main-]{paper}

% Authors

% \maketitle

% \thispagestyle{empty}

% \begin{figure*}[!htb]
%     \includegraphics[width=1\textwidth]{wacv-2026-author-kit-template/Figures/new_bio.pdf}
% \caption{Visualization of results for T2w and T1w Healthy brain MRI synthesis with corresponding error maps. McCaD yields lower artifacts with higher anatomical fidelity compared to baselines.} \label{qualitative_results_healthy}
% \end{figure*}

\begin{figure*}[!htb]
    \includegraphics[width=1\textwidth]{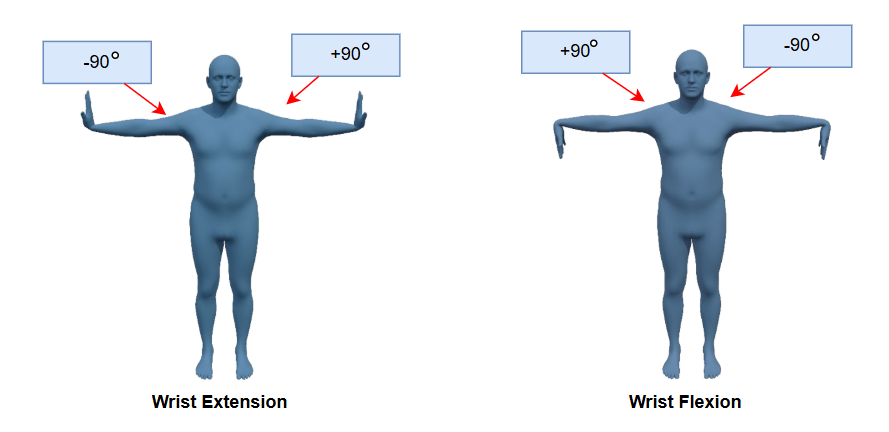}
\caption{We show extreme poses at $\pm 90^\circ$ for wrist extension/flexion sign convention with left and right sign mirroring consistent with SMPL-X euler angle setup.}
 \label{fig:wrist_movements}
\end{figure*}

\begin{figure*}[t]
  \centering
  \begin{subfigure}[t]{0.48\linewidth}
    \centering
    \includegraphics[width=\linewidth]{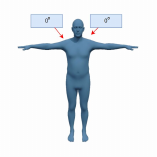}
    \caption{Maximum for horizontal abduction.}
    \label{fig:max_abd}
  \end{subfigure}\hfill
  \begin{subfigure}[t]{0.48\linewidth}
    \centering
    \includegraphics[width=\linewidth]{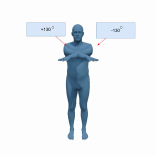}
    \caption{Maximum horizontal adduction (±130°).}
    \label{fig:max_add}
  \end{subfigure}
  \caption{Shoulder horizontal ad/abduction within signer space in SMPL-X compatible euler angle setup. Shoulder motion is constrained within a torso-anchored, ground-parallel signer space, disallowing horizontal abduction behind the torso while permitting substantial horizontal adduction for cross-body movements.}
  \label{fig:shoulder_constraints}
\end{figure*}

\begin{figure}[t]
    \centering
    \includegraphics[width=.4\textwidth]{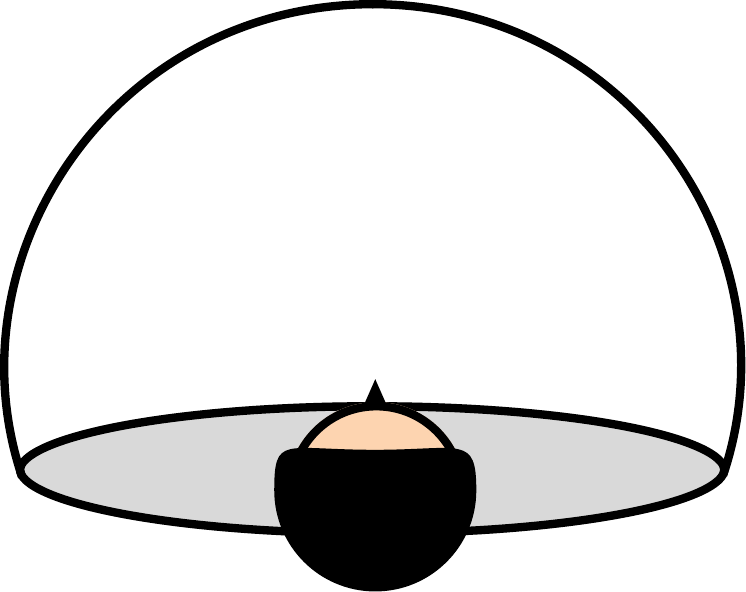}
    \caption{Bird's eye view of signer space envelope showing a torso-anchored 3D workspace. The figure has been adapted from~\cite{branchini2020grammar}.}
    \label{fig:signer_space}
\end{figure}

% \begin{figure*}[h!]
%     \centering
%     \begin{subfigure}[b]{0.48\textwidth}
%         \centering
%         \includegraphics[width=\textwidth]{wacv-2026-author-kit-template/Figures/1.pdf}
%         %\label{fig:a}
%     \end{subfigure}
%     \hfill
%     \begin{subfigure}[b]{0.48\textwidth}
%         \centering
%         \includegraphics[width=\textwidth]{wacv-2026-author-kit-template/Figures/3.pdf}
%         %\label{fig:b}
%     \end{subfigure}
%     \caption{Scenarios of participant engagement with the VAs (A) A participant interacting with the VA screen while browsing through images. (B) A group of participants engaging with the VA in a shared session.}
%     \label{fig:two_images}
% \end{figure*}

\section{Background Knowledge}
SMPL-X \cite{smplx} is an advanced parametric human body model, an extension of the original SMPL~\cite{loper2023smpl}, integrating hand articulations through MANO~\cite{romero2022embodied} and facial expressions through FLAME~\cite{li2017learning}.
This enables comprehensive full-body representations that include hand and face dynamics.
SMPL-X is defined by a mapping function $M(\boldsymbol{\theta}, \boldsymbol{\beta}, \boldsymbol{\psi}): \mathbb{R}^{|\boldsymbol{\theta}| \times|\boldsymbol{\beta}| \times|\boldsymbol{\psi|}} \rightarrow \mathbb{R}^{3 N}$, parameterized by the pose $\boldsymbol{\theta} \in \mathbb{R}^{3(K+1)}$, where $K$ is the number of body joints in addition to a joint for global rotation. $\boldsymbol{\beta}$ represents shape coefficients, and $\boldsymbol{\psi}$ are the facial expression coefficients.

The model uses vertex-based linear blend skinning with $N=10,475$ vertices and $K=54$ joints, including joints for hands, neck, jaw, and eyeballs.
The formulation of SMPL-X is defined as follows:

\begin{equation}
    M(\boldsymbol{\beta}, \boldsymbol{\theta}, \boldsymbol{\psi}) = W\left(T_P(\boldsymbol{\beta}, \boldsymbol{\theta}, \boldsymbol{\psi}), J(\boldsymbol{\beta)}, \boldsymbol{\theta}, \mathcal{W} \right),
\end{equation}
where,

\begin{equation}
\label{eq:smplx_template}
    \begin{split}
        T_P(\boldsymbol{\beta}, \boldsymbol{\theta}, \boldsymbol{\psi}) = \\
        \bar{T} + B_S(\boldsymbol{\beta}; \mathcal{{S}}) + B_E(\boldsymbol{\psi}; \mathcal{{E}}) + B_P(\boldsymbol{\theta}; \mathcal{{P}}).
    \end{split}
\end{equation}

$B_P(\cdot)$, $B_S(\cdot)$, and $B_E(\cdot)$ in Eq.~\eqref{eq:smplx_template} represent the pose, shape, and expression-dependent corrective blend functions.
$\mathcal{S}$, $\mathcal{E}$, and $\mathcal{P}$ represent the orthonormal principal components of vertex displacements of shape, pose, and expression blend shape variations.
These functions apply vertex displacements to the canonical template mesh $\bar{T}$ based on the pose parameters $\boldsymbol{\theta}$, shape parameters $\boldsymbol{\beta}$, and expression parameters $\boldsymbol{\psi}$.
In particular, $B_P(\boldsymbol{\theta})$ and $B_S(\boldsymbol{\beta})$ capture non-linear deformations specific to pose and shape variations.
After these corrections, the deformed mesh is processed using linear blend skinning, denoted as $\mathcal{W}$, which rotates the vertices around the joints $J(\boldsymbol{\beta})$ according to the skeletal kinematics.
The final mesh is smoothed using a predefined set of blend weights $W$, resulting in the articulated 3D human body mesh.

\section{Range of Motion and Signer Space}
This section complements section 3.2.1 of the main paper.
%We constrain the upper limb using the physiological degrees of freedom (DOFs)~\cite{hamill2006biomechanical,knudson2007fundamentals} of the joints most active in SL. 
We constrain the upper limb using the physiological degrees of freedom (DOFs)~\cite{hamill2006biomechanical,knudson2007fundamentals} of the joints most active in sign language. The shoulder has three DOFs, \ie, flexion/extension, abduction/adduction, and internal/external rotation. The elbow–forearm complex has two DOFs, \ie, humeroulnar flexion/extension and radioulnar pronation/supination. For the wrist, we adopt a three DOF formulation covering forearm pronation/supination, wrist flexion/extension, and radial/ulnar deviation. 

Clinical range of motion (ROM)~\cite{hamill2006biomechanical,knudson2007fundamentals} for these DOFs is typically reported as \emph{unsigned} magnitudes in anatomical planes (\eg, wrist flexion $90^\circ$ and extension $90^\circ$). 
%To use these values, we convert them to \emph{signed} bounds in an anatomical Euler~\cite {wikipedia_euler_angles} convention.
To use these values, we convert them to \emph{signed} bounds in an anatomical Euler~\cite{wikipedia_euler_angles} convention.
%For each DOF we align a local axis with the corresponding motion, adopt the right-hand rule for signs, and verify orientation on the rig. 
For each DOF, we align a local axis with the corresponding motion, adopt the right-hand rule for signs, and verify the orientation on the rig.

We express ROMs as a single signed interval in the aligned Euler convention. For bilateral joints, we mirror the sign across the sagittal plane (including wrist flexion/extension), matching the left/right labeling in Fig.~\ref{fig:wrist_movements}. This deterministic normalization yields SMPL-X~\cite{smplx} compatible signed bounds from clinical ROM with a brief visual sanity check.

Having normalized clinical ROM to SMPL-X compatible signed Euler bounds, we further constrain the shoulder using the notion of signer space~\cite{branchini2020grammar,wilcox2020conceptualization}, a torso-centric 3D region where signs are typically produced (see Fig.~\ref{fig:signer_space}). We model this as a compact volume anchored to the torso, bounded laterally near the shoulders, vertically from the lower chest to the forehead, and in depth slightly in front of the chest.

We restrict shoulder motion to be consistent with a torso-anchored, ground-parallel signer space envelope (see Fig.~\ref{fig:signer_space}). Accordingly, we cap horizontal abduction by disallowing motion behind the torso (see Fig.~\ref{fig:max_abd}) while permitting substantial horizontal adduction to support cross-body movements (see Fig.~\ref{fig:max_add}).
This yields a simple deterministic rule that filters out poses with shoulder angles outside these bounds.

\section{Motion Capture Data Acquisition}
 
\begin{figure}[t]
  \centering
  \includegraphics[width=\linewidth]{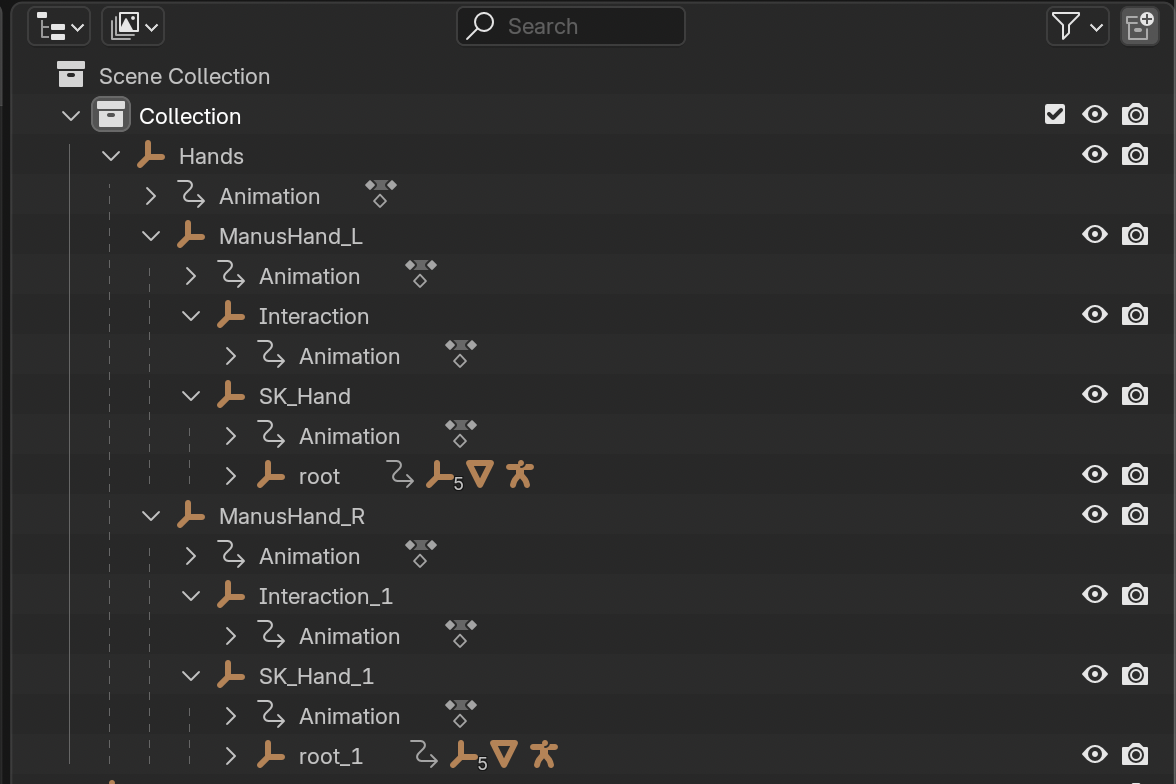}
  \caption{FBX hierarchy from the MANUS export. \texttt{Hands} controls global placement. \texttt{ManusHand\_L} and \texttt{ManusHand\_R} parent the \texttt{SK\_Hand} meshes and a per-hand \text{root} armature with finger and thumb mocap. \text{Animation} and \text{Interaction} store non-finger transforms that we remove for retargeting.}
  \label{fig:fbx-hierarchy}
\end{figure}

\begin{figure}[t]
  \centering
  \includegraphics[width=\linewidth]{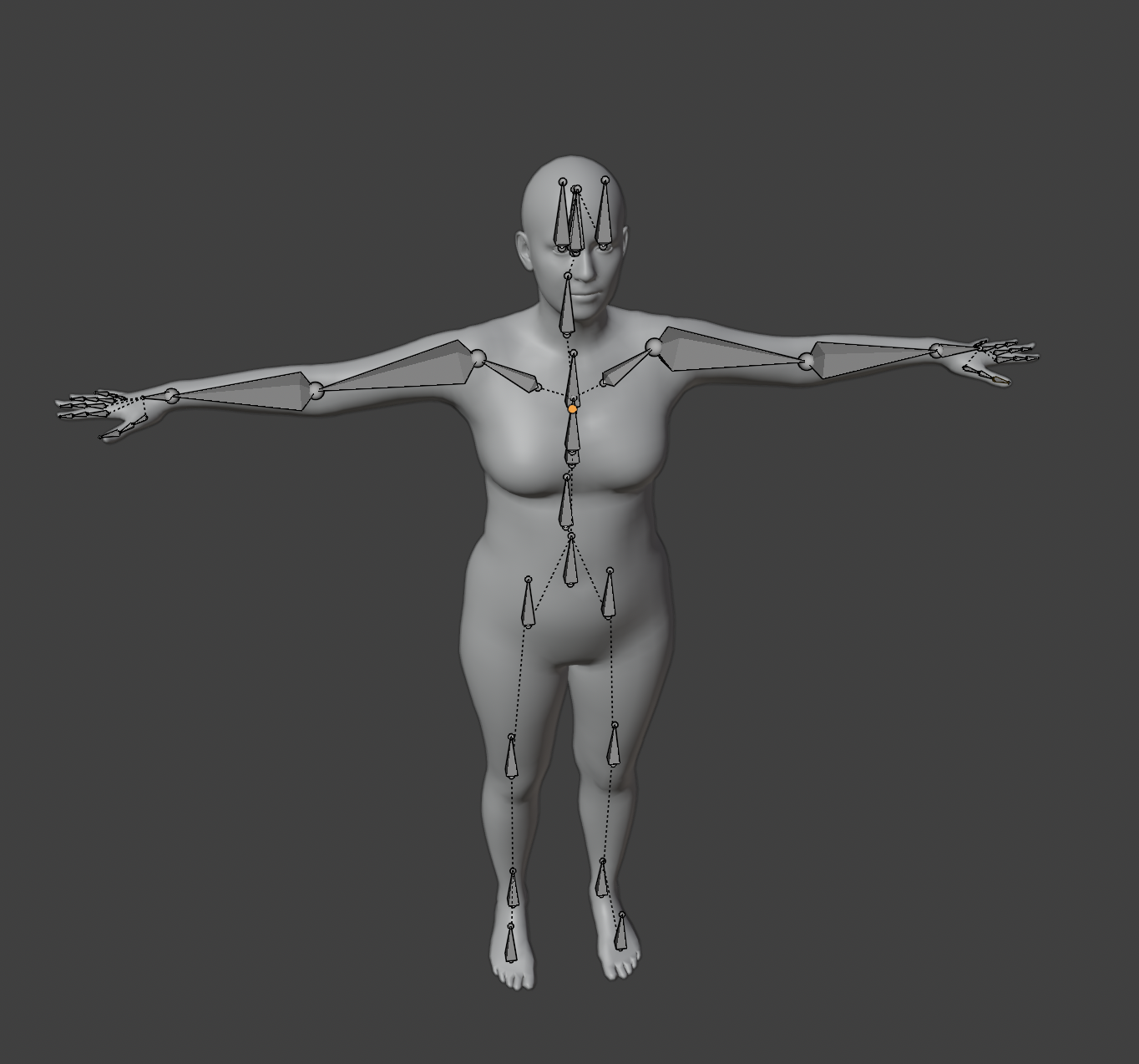}
  \caption{SMPL-X in the rest pose with visible armature.}
  \label{fig:smplx-tpose}
\end{figure}

% Figure reference used above: Fig.~\ref{fig:hand-armatures}
\begin{figure}[t]
  \centering
  \includegraphics[width=\linewidth]{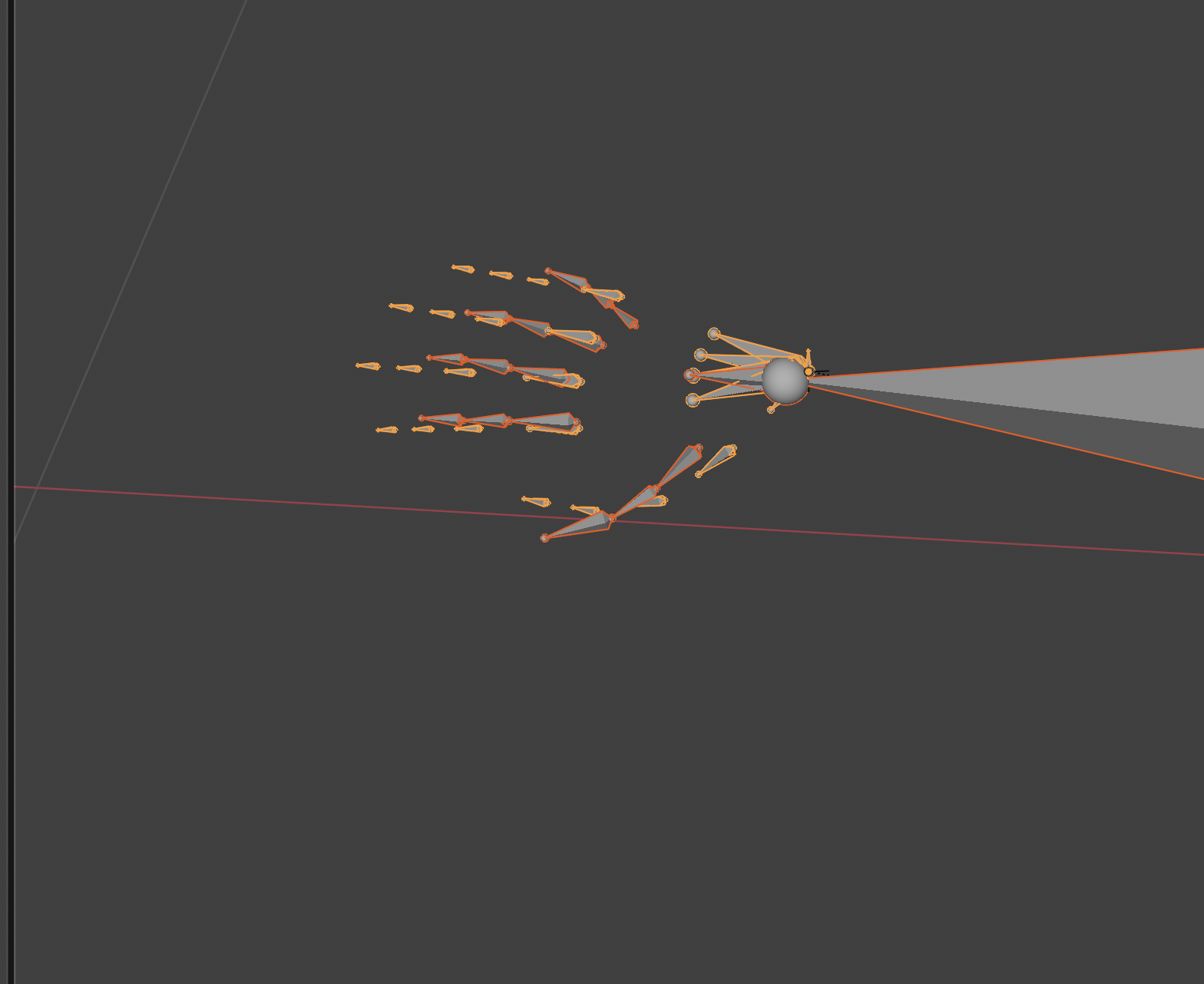}
  \caption{Separate hand armatures during retargeting. We retarget each hand one at a time. We duplicate the SMPL-X rig, retarget the right hand first, then retarget the left hand on the duplicate. We copy the left-hand keyframes from the duplicate back to the original SMPL-X rig, consolidating both hands' motion. Finger animations are then transferred to SMPL-X.}
  \label{fig:hand-armatures}
\end{figure}

This section complements section 3.2.2 of the main paper.
We export MANUS~\cite{manus_metagloves_pro} motion from Unity as FBX and process it in Blender. The FBX hierarchy in Fig.~\ref{fig:fbx-hierarchy} comprises a top-level Hands node that controls global translation, 
child nodes \texttt{ManusHand\_L} and \texttt{ManusHand\_R}, per-hand \texttt{SK\_Hand} meshes, and a terminal \texttt{root} that contains the armature. 

Rotations in mocap reside on the finger and thumb bones of the armature, while the parent nodes position the gloves about the wrist pivot.
After testing several approaches, the most reliable process was to set the armature to a flat ``rest'' pose, then delete keyframes on the parent containers so that only the finger animations remained. The armature was then aligned as closely as possible with the SMPL-X finger bones as shown in Fig.~\ref{fig:smplx-tpose}. Differences in finger length did not affect retargeting substantially, but misaligned knuckles caused distortion and stretching in the SMPL-X rig, so careful attention was paid to joint spacing. Some discrepancies remained due to restrictions on altering the SMPL-X T-pose. 

Since the hands were on separate armatures, they were retargeted one at a time as shown in Fig.~\ref{fig:hand-armatures}. We duplicated the SMPL-X rig, retargeted the right hand first, then retargeted the left hand on the duplicate. The left-hand keyframes from the duplicate were copied back to the original SMPL-X rig, consolidating both hands' motion. At this stage, the finger animations were successfully transferred to SMPL-X.
A fresh copy of the \texttt{.fbx} file was imported, and only the parent ``Hands'' keyframes were removed. This preserved the movement of the hands in space, allowing them to be positioned naturally in front of the SMPL-X body. 

To enable the arms to move with the MANUS gloves, an inverse kinematics (IK) setup was added. Each arm was given an IK handle constrained to a duplicate wrist bone, which was then linked to \texttt{ManusHand\_L} and \texttt{ManusHand\_R}. This enabled the arms to follow the hand movements using a single bone, rather than requiring manual adjustment of the forearm and upper arm. Wrist rotations could not be transferred due to incompatibilities between the SMPL-X T-pose and the bone roll of the MANUS rig, which meant that constraints could not replicate these rotations accurately. Finally, we baked the animations into the SMPL-X armature, replacing the constraints with explicit per-frame rotation keyframes. The temporary constraints and auxiliary IK bones were removed, leaving a clean animation that could be extracted and used.

\section{Analysis of Prior Parameters}

This section complements section 5.1 of the main paper.
We study how data filtering and lightweight bio-mechanical constraints affect body and hand pose estimation. Tables~\ref{tab:signbposer_variants} and~\ref{tab:signhposer_variants} report hyperparameter sweeps for SignBPoser and SignHPoser under matched architectures. Each setting varies only the training data correction and the presence of a bio-mechanical loss, while we select the best hyperparameter on Evaluation (DEV) and TEST data. We evaluate with MPJPE and MPVPE on both splits, and we summarize the main trends below.

\begin{table*}[t]
\centering
\caption{\textbf{Hyper-parameter tuning of SignBPoser}. We denote SignBPoser trained on $\text{BP}_{\text{u}}$: unfiltered body data, $\text{BP}_{\text{f}}$: filtered body data, $\text{BP}_{\text{f+bio}}$: filtered body data with body bio-mechanical loss. We evaluate using Mean Per Joint Position Error (MPJPE) and Mean Per Vertex Position Error (MPVPE), on the recovered joints and meshes.}
\label{tab:signbposer_variants}
\small
\begin{tabular}{ccccccccc}
\toprule\multirow{2}{*}{\textbf{Variant}} & \multicolumn{4}{c}{\textbf{Parameters}} & \multicolumn{2}{c}{\textbf{DEV}} & \multicolumn{2}{c}{\textbf{TEST}} \\
 & \textbf{KL} & \textbf{Neuron} & \textbf{Latent} & \textbf{Biomech constant} & \textbf{MPJPE$\downarrow$} & \textbf{MPVPE$\downarrow$} & \textbf{MPJPE$\downarrow$} & \textbf{MPVPE$\downarrow$} \\
\midrule
\multirow{3}{*}{$\text{BP}_{\text{u}}$} 
 & \textbf{0.001}  & \textbf{512} & \textbf{33} & \textcolor{red}{\xmark} & \textbf{5.87} & \textbf{3.73} & \textbf{5.69} & \textbf{3.62} \\
 & 0.001  & 512 & 32 & \textcolor{red}{\xmark} & 5.99 & 4.17 & 5.98  &  4.21 \\
 & 0.001  & 512 & 31 & \textcolor{red}{\xmark} & 7.56 & 5.05 & 7.28  & 5.00  \\
\midrule
\multirow{3}{*}{$\text{BP}_{\text{f}}$} 
 & \textbf{0.001}   & \textbf{512} & \textbf{33} & \textcolor{red}{\xmark} & \textbf{7.21} & \textbf{4.33} & \textbf{7.04} & \textbf{4.14}  \\
 & 0.001   & 512 & 32 & \textcolor{red}{\xmark} & 7.45 & 4.68 & 7.43 & 4.32 \\
 & 0.001   & 512 & 31 & \textcolor{red}{\xmark} & 7.37 & 4.41 & 7.17 & 4.24 \\
\midrule
\multirow{3}{*}{$\text{BP}_{\text{f+bio}}$} 
 & 0.001   & 512 & 33 & 0.5 & 7.42 & 4.43  & 7.21 & 4.32  \\
 & \textbf{0.001}   & \textbf{512} & \textbf{33} & \textbf{1.5} & \textbf{7.30}  & \textbf{4.39}  & \textbf{7.10}  & \textbf{4.25}  \\
 & 0.001   & 512 & 33 & 2.5 & 7.37 & 4.42 & 7.29 & 4.29 \\
\bottomrule\end{tabular}
\end{table*}

Table~\ref{tab:signbposer_variants} reports hyperparameter tuning for three settings of SignBPoser. $\text{BP}_{\text{u}}$ uses the unfiltered data, $\text{BP}_{\text{f}}$ uses the bio-mechanically filtered data, and $\text{BP}_{\text{f+bio}}$ adds a body bio-mechanical loss on top of the filtered data. We select the best hyperparameter for each setting on the DEV and TEST sets. 

In the unfiltered setting $\text{BP}_{\text{u}}$ latent 31 performs worst. Increasing to 32 reduces error on DEV by 21\% on MPJPE and 17\% on MPVPE, and on TEST by about 18\% and 16\%. A further increase to 33 brings additional reductions of about 2\% and 11\% on DEV, and about 5\% and 14\% on TEST. Latent 33 is therefore the most reliable choice in this setting.

On the other hand, in the bio-mechanical filtered setting $\text{BP}_{\text{f}}$, latent 32 has the highest error. Switching to 31 reduces error on DEV by about 1\% MPJPE and 6\% MPVPE, and on TEST by about 3\% and 2\%. Increasing to 33 brings smaller additional gains of roughly 2\% on both metrics on DEV and about 2\% on TEST. The effect of latent size is therefore mild in this setting.

Finally for filtered-plus-constraint setting $\text{BP}_{\text{f+bio}}$, the extremes $0.5$ and $2.5$ give slightly higher errors. Setting the weight to $1.5$ reduces MPJPE by about $1\text{--}3\%$ and MPVPE by about $1\%$ on both DEV and TEST. The configurations are close to each other, which indicates stable behavior with the constraint.

\begin{table*}[t]
\centering
\caption{\textbf{Hyper-parameter tuning of SignHPoser}. We denote SignHPoser trained on $\text{HP}_{\text{u}}$: unfiltered hand data, $\text{HP}_{\text{f}}$: filtered hand data, $\text{HP}_{\text{f+bio}}$: filtered hand data with hand bio-mechanical loss.}
\label{tab:signhposer_variants}
\small
\begin{tabular}{ccccccccc}
\toprule\multirow{2}{*}{\textbf{Variant}} & \multicolumn{4}{c}{\textbf{Parameters}} & \multicolumn{2}{c}{\textbf{DEV}} & \multicolumn{2}{c}{\textbf{TEST}} \\
 & \textbf{KL} & \textbf{Neuron} & \textbf{Latent} & \textbf{Biomech constant} & \textbf{MPJPE$\downarrow$} & \textbf{MPVPE$\downarrow$} & \textbf{MPJPE$\downarrow$} & \textbf{MPVPE$\downarrow$} \\
\midrule
\multirow{3}{*}{$\text{HP}_{\text{u}}$} 
 & \textbf{0.0001}  & \textbf{512} & \textbf{24} & \textcolor{red}{\xmark} & \textbf{0.56} & \textbf{0.55} & \textbf{0.56} & \textbf{0.54} \\
 & 0.0001  & 512 & 23 & \textcolor{red}{\xmark} & 0.57 & 0.58 & 0.57 & 0.55 \\
 & 0.0001  & 512 & 22 & \textcolor{red}{\xmark} & 0.59 & 0.58 & 0.60 & 0.58 \\
\midrule
\multirow{3}{*}{$\text{HP}_{\text{f}}$} 
 & 0.0001   & 512 & 24 & \textcolor{red}{\xmark} & 0.40 & 0.38 & 0.40 & 0.38 \\
  & \textbf{0.0001}   & \textbf{512} & \textbf{23} & \textcolor{red}{\xmark} & \textbf{0.37} & \textbf{0.35} & \textbf{0.37} & \textbf{0.35} \\
 & 0.0001   & 512 & 22 & \textcolor{red}{\xmark} & 0.38 & 0.40 & 0.38 & 0.40 \\
\midrule
\multirow{3}{*}{$\text{HP}_{\text{f+bio}}$} 
 & 0.0001   & 512 & 23 & 0.5 & 0.40 & 0.41 & 0.40 & 0.41 \\
 & \textbf{0.0001}   & \textbf{512} & \textbf{23} & \textbf{1.5} & \textbf{0.39} & \textbf{0.38} & \textbf{0.39} & \textbf{0.38} \\
 & 0.0001   & 512 & 23 & 2.5 & 0.43 & 0.45 & 0.43 & 0.45 \\
\bottomrule
\end{tabular}
\end{table*}

In Table~\ref{tab:signhposer_variants} we compare three settings for SignHPoser. $\text{HP}_{\text{u}}$ is trained on the uncorrected data, $\text{HP}_{\text{f}}$ is trained on a bio-mechanically corrected data, and $\text{HP}_{\text{f+bio}}$ keeps the corrected poses and adds a lightweight hand bio-mechanical loss during training. The architecture matches SignBPoser, and the only differences are data correction and the presence of the hand constraint. 

In the $\text{HP}_{\text{u}}$ setting, moving from latent 22 to 23 lowers error by about 5\% on MPJPE and about 5\% on MPVPE on TEST, with similar gains on DEV. Increasing to 24 gives an additional reduction of roughly 2\% on MPJPE across splits, and about 2\% on MPVPE on TEST and about 5\% on DEV. Latent 24 is therefore the most reliable within this setting.

For $\text{HP}_{\text{f}}$, both latent 24 and 22 are worse than 23. From 24 to 23 the error drops by about 8\% on MPJPE and about 8\% on MPVPE on both DEV and TEST. From 22 to 23, the drop is smaller on MPJPE at about 3\%, but larger on MPVPE at about 12–14\% across splits. Latent 23 is the preferred choice.

Finally for $\text{HP}_{\text{f+bio}}$, weight 2.5 performs worst. Reducing it to 1.5 lowers MPJPE by about 9–10\% and MPVPE by about 16–18\% on both DEV and TEST. Compared with 0.5, the 1.5 setting also improves by about 2–3\% on MPJPE and about 8\% on MPVPE. The hand constraint is most effective at 1.5.

From the above results, we can conclude that training for SignBPoser and SignHPoser remains stable under a similar architecture. Simple choices like latent size and lightweight bio-mechanical constraints guide accuracy without instability across the DEV and TEST sets.

\section{Ablation of SignHPoser with Vposer}

\begin{table}[t]
    \setlength{\tabcolsep}{2pt}
    \centering
    \caption{Ablation study of SignHPoser with Vposer within \mymethod.}
    \label{tab:signhposer_variants_vposer}
    \begin{tabular}{lccc}
        \toprule
        \textbf{Method} & \textbf{UBody (-F)}$\downarrow$ & \textbf{LHand}$\downarrow$ & \textbf{RHand}$\downarrow$ \\
        \midrule
        $\text{HP}_{\text{u}}$ & 37.25 & 13.56 & 14.53 \\
        $\text{HP}_{\text{f}}$ & 36.79 & 13.39 & 14.06 \\
        $\text{HP}_{\text{f+bio}}$ & \textbf{36.77} & \textbf{13.37} & \textbf{13.82} \\
        \bottomrule
    \end{tabular}
\end{table}

This section complements section 5.1 of the main paper. We evaluate the hand prior SignHPoser using VPoser. Table~\ref{tab:signhposer_variants_vposer} summarizes ablation results.
The first two rows of Table~\ref{tab:signhposer_variants_vposer} show the results for $\text{HP}_{\text{u}}$ and $\text{HP}_{\text{f}}$ variants.
It can be observed that $\text{HP}_{\text{f}}$ outperforms $\text{HP}_{\text{u}}$ on all metrics, with relative error reductions of \textbf{1.2\%} on Upper Body, \textbf{1.3\%} on Left Hand, and \textbf{3.2\%} on Right Hand. This demonstrates the importance of the correction process using bio-mechanical constraints on the hand data.
Adding a bio-mechanical regularizer to the filtered prior in $\text{HP}_{\text{f+bio}}$ increases accuracy compared to $\text{HP}_{\text{f}}$, yielding slight improvements on Upper Body (0.05\%), Left Hand (0.15\%), and Right Hand (1.7\%).
This indicates that introducing bio-mechanical constraints provides useful physical regularization in the fitting process.

\section{Limitations of the SGNify Ground Truth}
\begin{figure*}[t]
  \centering
  \includegraphics[width=0.9\textwidth]{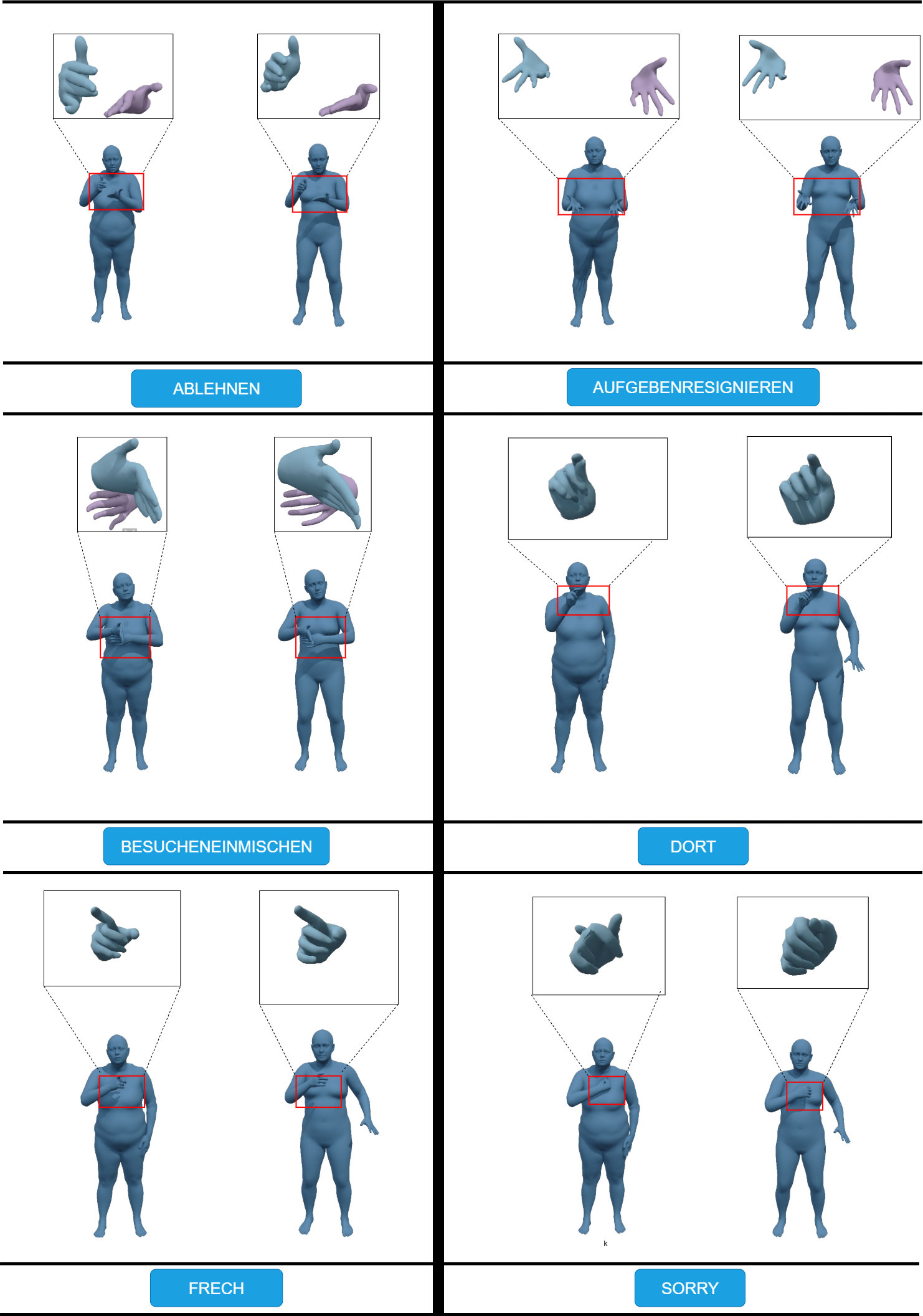}
  \caption{\textbf{Examples of independent signs}. Each panel shows two signers, the left signer is the ground truth from the SGNiFy evaluation set and the right signer is DexAvatar generated by our fitting optimization. The SGNiFy~\cite{forte2023reconstructing} ground truth often contains low-quality hand shapes and placements, while DexAvatar produces more plausible poses. The improvement comes from a SignHPrior trained on our mocap dataset.}
  \label{qualitative_results}
\end{figure*}

This section complements section 5.2 of the main paper.
%We evaluate with TR-V2V against SGNify ground truth, which contains occasional implausible hand configurations.
We evaluate using TR-V2V against the SGNify ground truth, which contains occasional implausible hand configurations.
%DexAvatar shows consistent improvements on both hands and the upper body.
DexAvatar shows consistent improvements for both hands and the upper body.
%Improvements are consistent across one-handed and two-handed signs, indicating robustness to interaction asymmetry. The margin remains modest as TR-V2V penalizes distance to the ground truth mesh, and when the reference encodes anatomically inconsistent finger postures or knuckle spacing, moving toward a plausible pose does not always reduce vertex distance. 
The margin remains modest because TR-V2V penalizes distance to the ground truth mesh. When the reference encodes anatomically inconsistent finger postures or knuckle spacing, moving toward a plausible pose does not always reduce vertex distance.
%Ground truth often contain collapsed fingers and irregular knuckle spacing, which explains why plausibility corrections may not yield large vertex reductions. 
The ground truth often contains collapsed fingers and irregular knuckle spacing, which explains why plausibility corrections may not yield large vertex reductions.
Qualitative comparisons in Fig.~\ref{qualitative_results} show cleaner finger alignment. 
%For certain signs like BESUCHENEINMISCHEN and FRECH, \mymethod generates more plausible hand poses as compared to the ground truth. 
For certain signs like BESUCHENEINMISCHEN and FRECH, \mymethod{} generates more plausible hand poses compared to the ground truth.
These effects arise as SignHPoser optimizes toward anatomically plausible hand poses learned from our mocap data.

\section{Additional qualitative evaluations with SOTA method}

This section complements section 5.2 of the main paper. In Fig.~\ref{fig:qualitative_blur}, ~\ref{fig:qualitative_occ}, and ~\ref{fig:qualitative_gauss} we present qualitative results of \mymethod{} on the MM-WLAuslan~\cite{shen2024mm} SL dataset, compared against SGNify~\cite{forte2023reconstructing} and EVA* under challenging scenarios such as motion blur, self-occlusion, and noise.

\subsection{Qualitative evaluation on motion blur images}

\begin{figure*}[t]
  \centering
  \includegraphics[width=0.9\textwidth]{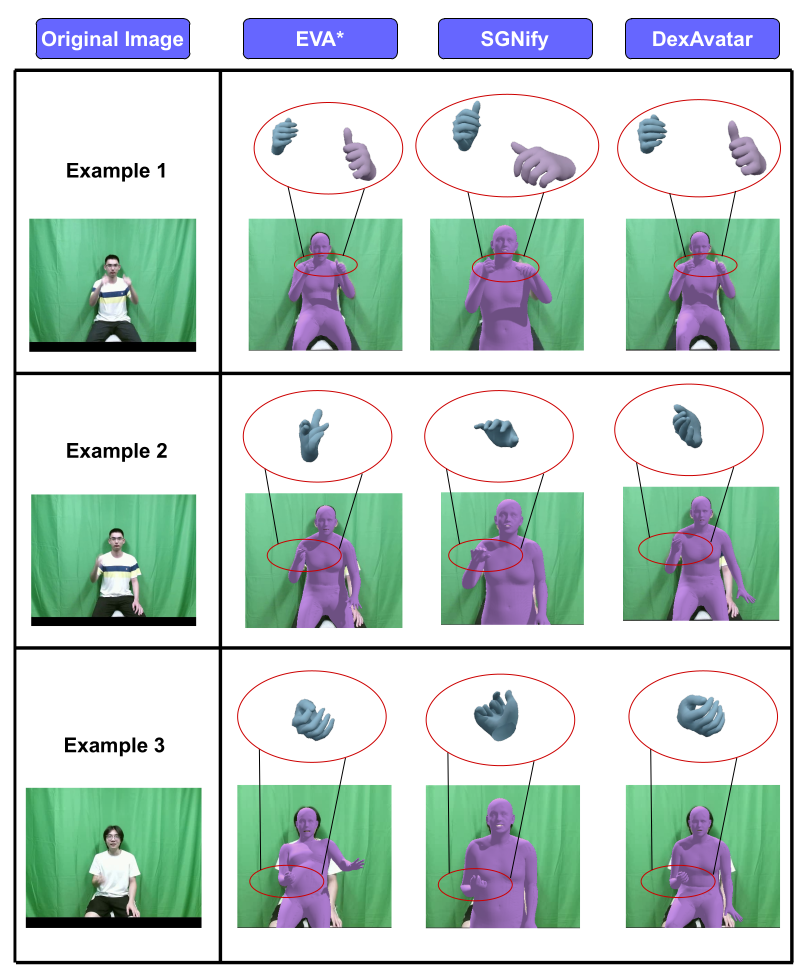}
  \caption{\textbf{Qualitative evaluation under motion blur}. We compare EVA*, SGNify~\cite{forte2023reconstructing}, and \mymethod{}. \mymethod{} preserves compact rounded finger configurations and clean contact, while EVA* overspreads or distorts the fingers and SGNify yields incorrect detections or misrepresented contact, with additional body misalignment appearing in harder cases.}
  \label{fig:qualitative_blur}
\end{figure*}

Fig.~\ref{fig:qualitative_blur} presents three examples under motion blur. In \texttt{Example 1}, EVA* shows overspread fingers with a unnatural bending and uneven gaps, while SGNify is an incorrect wrist configuration although bio-mechanically correct, whereas \mymethod{} maintains a compact rounded configuration with evenly spaced fingertips that matches the blurred target contact. Building on this, in \texttt{Example 2}, EVA* produces distorted fingers with incorrect spacing and no bio-mechanical stability, SGNify is again an incorrect detection although bio-mechanically correct, while \mymethod{} preserves clear fingers with realistic curl and fingertip positions that support the intended overlap with a clean stable contact. Continuing this pattern, in \texttt{Example 3}, EVA* compresses the fingers into a tight bundle so separation is lost and local interpenetration appears along the contact between the index finger and thumb, and also shows a body misalignment relative to the image, SGNify exhibits penetration of fingers with palm that are inconsistent with the image, whereas \mymethod{} forms a coherent rounded cluster with visible fingertip order and contact that follows the blurred evidence without fusion and is correct. Overall, \mymethod{} maintains accurate and plausible hand contacts under blur.

\subsection{Qualitative evaluation on self-occluded images}

\begin{figure*}[t]
  \centering
  \includegraphics[width=0.9\textwidth]{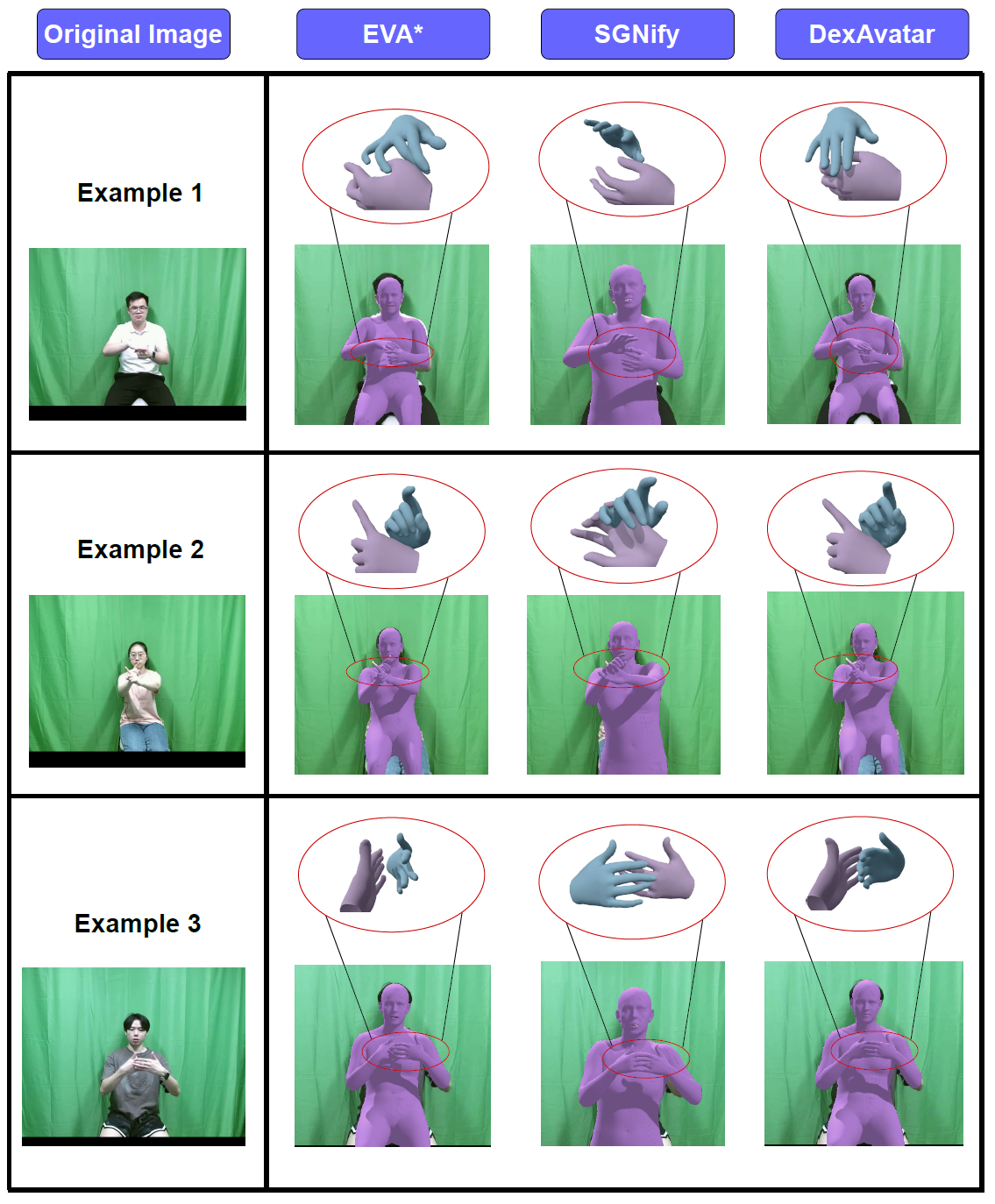}
  \caption{\textbf{Qualitative evaluation under self-occlusion}. We compare EVA*, SGNify~\cite{forte2023reconstructing}, and \mymethod{}. \mymethod{} maintains compact finger configurations and correct overlap, while EVA* overspreads or distorts the fingers and SGNify misplaces contact or infers the overlap order incorrectly.}
  \label{fig:qualitative_occ}
\end{figure*}

Fig.~\ref{fig:qualitative_occ} presents three examples under self-occlusion. In \texttt{Example 1}, EVA* overspreads the fingers with interpenetration of right hand with the left and inconsistent fingertip spacing, SGNify is bio-mechanically reasonable but misplaces the hand contact, whereas \mymethod{} maintains a compact closed configuration with fingertips aligned to the intended contact. Similarly, in \texttt{Example 2}, EVA* keeps both index fingers unnaturally extended while the others curl, SGNify reconstruction consists of overlapping of hands with a wrong wrist orientation for the left hand, while \mymethod{} preserves clear fingers with realistic curl and a clean stable contact. Finally, in \texttt{Example 3}, EVA* shows right hand finger postures beyond plausibility, SGNify is plausible but infers the overlap order incorrectly since the right hand should occlude the left rather than the reverse, whereas \mymethod{} provides the correct estimation with plausible hands throughout. Overall, \mymethod{} maintains accurate and plausible hand contacts under occlusion.

\subsection{Qualitative evaluation on images with gaussian noise}

\begin{figure*}[t]
  \centering
  \includegraphics[width=0.9\textwidth]{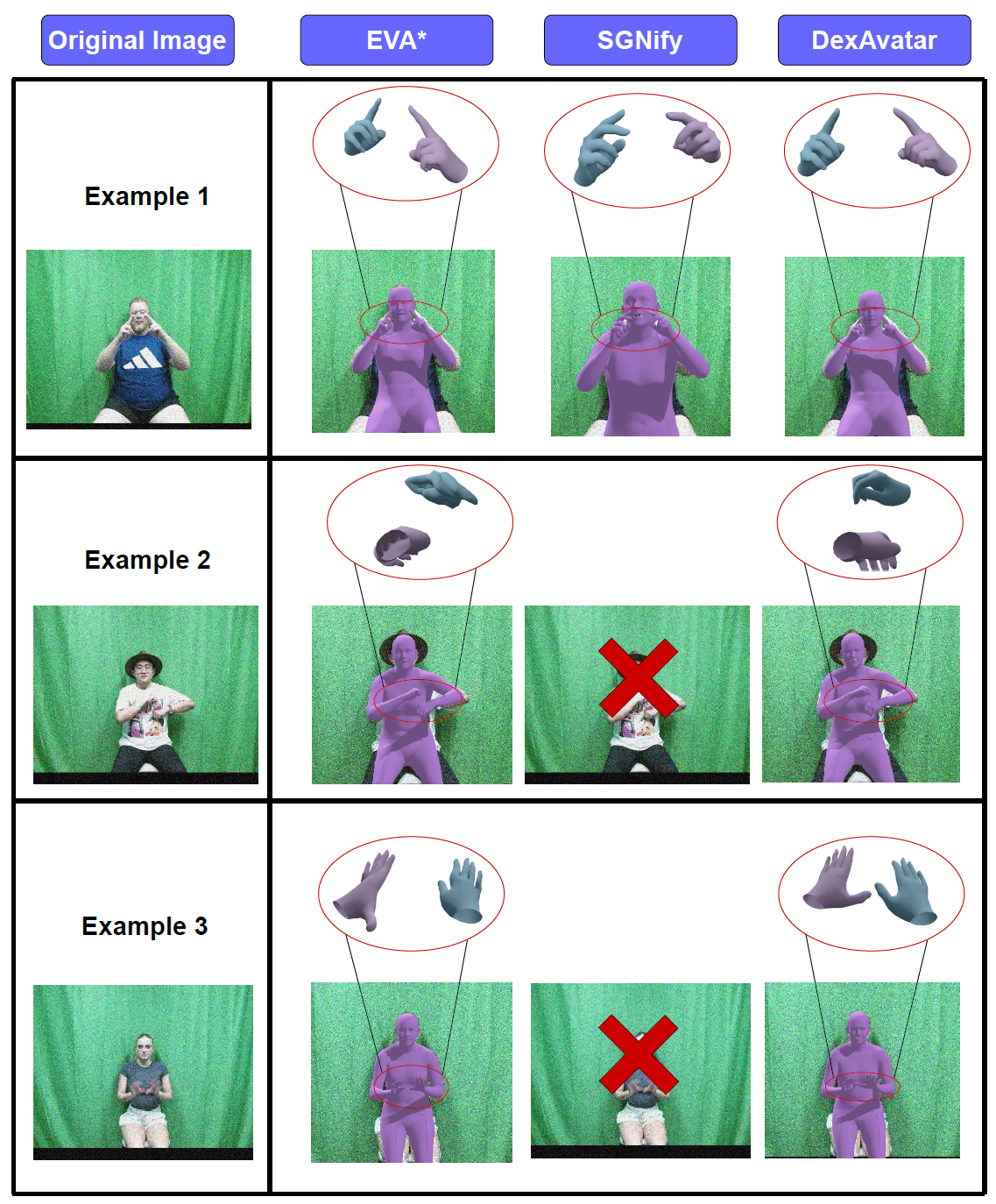}
  \caption{\textbf{Qualitative evaluation under gaussian noise}. We add Gaussian noise to input frames and compare EVA*, SGNify~\cite{forte2023reconstructing}, and \mymethod{}. \mymethod{} preserves plausible finger shape and clean contact, while EVA* exhibits implausible spacing and distortions, and SGNify fails to produce a mesh in harder cases due to missing keypoints.}
  \label{fig:qualitative_gauss}
\end{figure*}

We add Gaussian noise to the input frames and compare EVA*, SGNify~\cite{forte2023reconstructing}, and \mymethod{} in Fig.~\ref{fig:qualitative_gauss}. In \texttt{Example 1}, EVA* reconstructs excessive spacing between the pinky and ring fingers which is bio-mechanically implausible, while SGNify yields a bio-mechanically reasonable hand that nevertheless does not match the target configuration in the image. In contrast, \mymethod{} recovers the intended finger arrangement and maintains bio-mechanical constraints across both hands. Moving to \texttt{Example 2}, EVA* degrades to an implausible finger arrangement and SGNify fails to produce any mesh due to missing keypoints under noise, whereas \mymethod{} still reconstructs a complete and accurate pose with plausible finger angles and stable contact. Finally, in \texttt{Example 3}, EVA* shows a left-thumb posture beyond plausibility together with distorted right-hand fingers and SGNify again produces no mesh because no keypoints are detected, while \mymethod{} returns an anatomically plausible reconstruction with unbroken fingers and consistent bilateral alignment. Overall, \mymethod{} remains stable under noisy frames and preserves both accuracy and bio-mechanical plausibility.

\clearpage
\twocolumn  % Force two-column mode to restart properly
%{\small
%\bibliographystyle{ieeenat_fullname}
%\bibliography{supp}
%}

\newpage
{
    \small
    \bibliographystyle{ieeenat_fullname}
    \bibliography{main}
}

\end{document}